\documentclass[10pt,journal,compsoc]{IEEEtran}
\usepackage[american]{babel}
\usepackage{amsmath,amsfonts}
\usepackage{algorithmic}
\usepackage{algorithm}
\usepackage{array}
\usepackage[caption=false,font=normalsize,labelfont=sf,textfont=sf]{subfig}
\usepackage{textcomp}
\usepackage{stfloats}
\usepackage{url}
\usepackage{verbatim}
\usepackage{graphicx}
\usepackage{cite}
\usepackage{enumitem}
\usepackage{xcolor}
\usepackage{makecell}
\usepackage{adjustbox}
\usepackage{multirow}
\usepackage{threeparttable}
\usepackage{colortbl}
\usepackage{color}
\usepackage{amssymb}
\usepackage{gensymb}
\usepackage{textcomp}
\usepackage{booktabs}
\usepackage{float}

\usepackage[colorlinks,bookmarksnumbered,bookmarksopen,linkcolor=black,citecolor=black,urlcolor=black]{hyperref}

\usepackage{ragged2e}
\renewcommand{\raggedright}{\leftskip=0pt \rightskip=0pt plus 0cm}

\makeatletter

\newcommand{\Rmnum}[1]{\expandafter\@slowromancap\romannumeral #1@}
\makeatother

\begin{document}

\title{Neovascularization Segmentation via a Multilateral Interaction-Enhanced Graph Convolutional Network}

\author{Tao Chen, Dan Zhang, Da Chen, Huazhu Fu, Kai Jin, Shanshan Wang,\\ Laurent D. Cohen, Yitian Zhao, Quanyong Yi, Jiong Zhang

\thanks{T. Chen, Y. Zhao, J. Zhang are with the Laboratory of Advanced Theranostic Materials and Technology, Ningbo Institute of Materials Technology and Engineering, Chinese Academy of Sciences, Ningbo, China. (jiong.zhang@ieee.org)}
\thanks{D. Zhang is with the School of Cyber Science and Engineering, Ningbo University of Technology, Ningbo, China.}
\thanks{D. Chen is with the Shandong Artificial Intelligence Institute, Qilu University of Technology, Shandong Academy of Sciences, Shandong, China.}
\thanks{H. Fu is with the Institute of High Performance Computing, A*STAR, Singapore.}
\thanks{K. Jin is with the Eye Center, The Second Affiliated Hospital, Zhejiang University, Hangzhou, China}
\thanks{S. Wang is with the Shenzhen Institute of Advanced Technology, Chinese Academy of Science, Shenzhen, China}
\thanks{Laurent D. Cohen is with the University Paris Dauphine, PSL Research University, CNRS, UMR 7534, CEREMADE, 75016 Paris, France.}
\thanks{Q. Yi is with the Ningbo Eye Hospital, Wenzhou Medical University, Ningbo, Zhejiang, China.}
}


\IEEEtitleabstractindextext{
\begin{abstract}
\raggedright{
Choroidal neovascularization (CNV), a primary characteristic of wet age-related macular degeneration (wet AMD), represents a leading cause of blindness worldwide. In clinical practice, optical coherence tomography angiography (OCTA) is commonly used for studying CNV-related pathological changes, due to its micron-level resolution and non-invasive nature. Thus, accurate segmentation of CNV regions and vessels in OCTA images is crucial for clinical assessment of wet AMD. However, challenges existed due to irregular CNV shapes and imaging limitations like projection artifacts, noises and boundary blurring. Moreover, the lack of publicly available datasets constraints the CNV analysis. To address these challenges, this paper constructs the first publicly accessible CNV dataset (\textbf{CNVSeg}), and proposes a novel multilateral graph convolutional interaction-enhanced CNV segmentation network (\textbf{MTG-Net}). This network integrates both region and vessel morphological information, exploring semantic and geometric duality constraints within the graph domain. Specifically, MTG-Net consists of a multi-task framework and two graph-based cross-task modules: Multilateral Interaction Graph Reasoning (MIGR) and Multilateral Reinforcement Graph Reasoning (MRGR). The multi-task framework encodes rich geometric features of lesion shapes and surfaces, decoupling the image into three task-specific feature maps. MIGR and MRGR iteratively reason about higher-order relationships across tasks through a graph mechanism, enabling complementary optimization for task-specific objectives. Additionally, an uncertainty-weighted loss is proposed to mitigate the impact of artifacts and noise on segmentation accuracy. 
Experimental results demonstrate that MTG-Net outperforms existing methods, achieving a Dice socre of 87.21\% for region segmentation and 88.12\% for vessel segmentation.}
\end{abstract}

}

\maketitle

\section{Introduction}
\begin{figure}[t]
\centering{
\includegraphics[width=8.5cm]{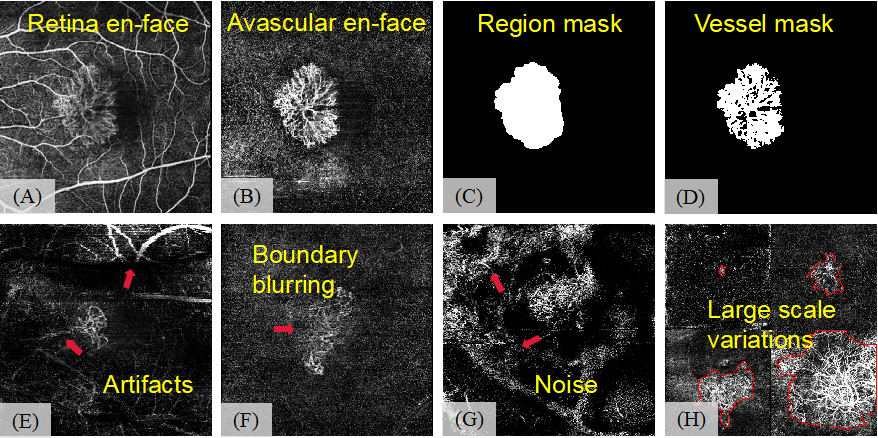}
}


\caption{The first row shows the en-face images of the retinal layer with CNV and the avascular layer collected by the OCTA device, as well as the lesion areas and vessels segmented by the automatic method (presented with labels). The second row illustrates the challenges faced in CNV segmentation.}
\label{intro_gcn}
\end{figure}

\IEEEPARstart{W}{et} age-related macular degeneration (wet AMD) is one of the leading causes of blindness worldwide~\cite{nowak2006age}, primarily attributed to choroidal neovascularization (CNV)~\cite{grossniklaus2004choroidal}. The rupture of CNV and abnormal permeability cause varying degrees of visual impairment~\cite{su2012adverse}. In clinical practice, optical coherence tomography angiography (OCTA)~\cite{de2015review}~\cite{chalam2016optical}~\cite{zhang20193d} has emerged as a promising alternative for observing CNV, offering non-invasive, rapid acquisition, and micrometer-level resolution, as shown in  Fig.~\ref{intro_gcn}, (A) is the en-face image of the retinal layer collected by OCTA, and (B) is the en-face image of the avascular layer. However, the manual analysis of CNV pathology by ophthalmologists is both labor-intensive and costly. Currently, automatic image analysis methods based on deep learning have been widely used in natural and medical images~\cite{qiu2023rethinking}~\cite{wu2023querying}. Therefore, designing a deep learning CNV automatic extraction approach based on OCTA images is crucial for accurate and effective diagnosis of wet AMD. Fig.~\ref{intro_gcn} (C) and (D) show the automatic segmentation results, where the CNV lesion areas and vessel regions are presented using labels.


Currently, several deep learning-based techniques~\cite{xue2018automatic,feng2023automated,liu2015automated,wang2020automated,chen2023rbgnet,vali2023cnv,wang2023deep} have been developed for CNV segmentation, offering quantitative basis for clinical analysis. However, these methods still struggle to accurately segment CNV lesions and vessels with fine shapes and high accuracy. We believe this limitation is mainly due to image quality and disease characteristics, which make accurate CNV detection challenging in complex cases. As shown in (E)-(G) of Fig.~\ref{intro_gcn}, imaging noise and artifacts projected from the top retinal vessels to the bottom choroidal cause blurred CNV marginal areas with low visibility. Difficulty in distinguishing artifacts from lesion signals can lead to over-segmentation. In addition, as shown in (H) of Fig.~\ref{intro_gcn}, the complex, irregular CNV patterns and variations in lesion size pose challenges to algorithm accuracy, leading to performance decline. Overcoming these issues requires enhancing the model’s ability to detect CNV boundaries and shapes, enabling better distinction between lesions and artifacts.

To address the challenges in CNV segmentation, we propose a novel Multi-Task Graph Convolutional Segmentation Network (MTG-Net) based on feature space graph modeling. Our method uses semantic clustering to extract key region nodes and connects structurally stable areas to build an explicit graph. By leveraging the global information propagation and reasoning abilities of graph neural networks, MTG-Net effectively integrates region, boundary, and shape cues. This improves the model’s ability to handle CNV scale variations and provides a more complete representation of lesion shape, size, and spatial distribution. In contrast, existing methods based on local feature similarity~\cite{wang2018interactive} or pixel-level attention~\cite{chen2018encoder,oktay2018attention} struggle to capture the global structure of complex lesions and are more vulnerable to speckle noise and imaging artifacts. The explicit definition of nodes and edges in our graph structure allows for interpretable modeling of inter-region relationships, with clearer and more controllable information propagation paths, significantly improving the interpretability of the model.

Specifically, we introduce two new modules: Multilateral Interaction Graph Reasoning (MIGR) and Multilateral Reinforcement Graph Reasoning (MRGR). The MIGR enables collaborative reasoning among constructed multi-task graphs, allowing the model to better learn the region, boundary and shape information of CNVs. Meanwhile, the MRGR focuses on improving the visibility of boundaries and shapes in the region segmentation task. 
By constructing multiple graphs, our model can address the problems of domain-specific feature update and reasoning  at multiple levels, enhancing the interaction and reinforcement of information across tasks. This method effectively tackles the challenges posed by large scale variations in CNV and the complexities of learning boundaries and shapes, thereby enhancing the model's generalizability. Additionally, we introduce an uncertainty estimation loss. This loss reduces the impact of artifacts and noise by assigning higher weights to pixels with blurred boundaries, as determined by an uncertainty map generated during training. Overall, the main contributions of this work are fourfold:

\begin{enumerate}[left=1em]
\item[$\bullet$] We establish the first publicly accessible CNV segmentation dataset using OCTA images. This dataset, named CNVSeg, includes precise pixel-level annotations of regions and vessels from 184 wet AMD cases, captured by three widely used OCTA devices (Heidelberg, Vision Micro, and Zeiss). To facilitate further developments, we provide the datasets, code, and baseline models on this website~\footnote{\url{https://github.com/jiongzhang-john/CNV}}.

\item[$\bullet$] We develop a graph-based multi-task segmentation method, named MTG-Net, which is able to effectively guide the information transfer between different tasks and to leverage inter-task data to enhance CNV segmentation performance.

\item[$\bullet$] We design two novel modules, MIGR and MRGR, which utilize duality constraints between tasks to extract both intra- and inter-task information. This approach enhances the underlying representation for region segmentation.

\item[$\bullet$] We propose a new uncertainty-weighted loss to address the issue of blurred pixels at lesion boundaries caused by artifacts and noise. Simultaneously, the weights between losses are adaptively adjusted during training through uncertainty estimation, effectively enhancing the model's training efficiency.
\end{enumerate}

\begin{figure*}[!h]
\centering{
\includegraphics[width=0.9\linewidth]{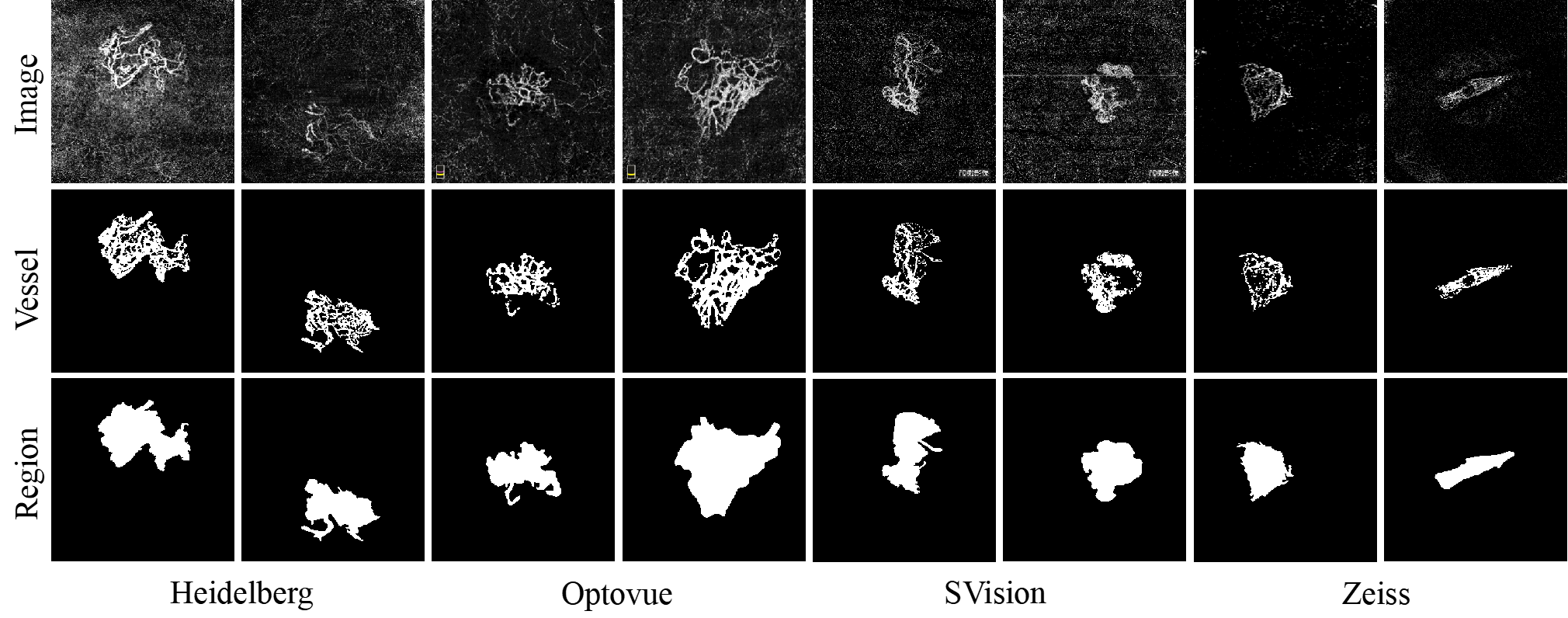}
}
\caption{The CNV dataset samples include data from four devices. The first row contains the original images, the second row contains the region labels, and the third row contains the vessel labels.}
\label{dataset_description}
\end{figure*}

\section{Related Work}
\subsection{CNV segmentation}
Previously, segmentation methods for CNV were primarily based on OCT B-scan images, aimed at distinguishing different types of CNV~\cite{xi2020ia,meng2021mf,zhang2019mpb,xi2019automated}. In their work, Zhang~\textit{et al.}~\cite{zhang2019mpb} employed multi-scale parallel branching and added supervision at the end of the branching to enhance the discriminative nature of feature representations and optimize the network. Meng~\textit{et al.}~\cite{meng2021mf} designed a multi-scale adaptive perceptual deformation module that directs the network to focus on the deformation of the target and aggregates contextual information. However, in some complex cases, structural OCT data cannot distinguish CNV from other lesions that cause retinal layer disruption, which affects segmentation accuracy.

With the increasing demands of clinical applications, and in order to better assist in the observation of pre-op and post-op treatment conditions of CNV vessels, as well as effectively handle complex cases, several methods have been designed based on OCTA en-face images~\cite{xue2018automatic,feng2023automated,liu2015automated,wang2020automated,chen2023rbgnet,vali2023cnv,wang2023deep}. These methods can be categorized into two types: i) traditional image detection algorithms: Liu~\textit{et al.}~\cite{liu2015automated} proposed a saliency-based algorithm to extract the CNV by using the signal strength and direction information. Xue~\textit{et al.}~\cite{xue2018automatic} proposed an unsupervised CNV segmentation method based on clustering algorithm called the density cell type P system. However, their approach still suffers from mis-segmentation when artifacts are indistinguishable from lesion signals. ii) deep learning method: Feng~\textit{et al.}~\cite{feng2023automated} achieved CNV segmentation by enhancing the traditional UNet with ResNet modules and spatial pyramid pooling modules. Chen~\textit{et al.}~\cite{chen2023rbgnet} designed a combination of a transformer and a CNN to solve the problem of the large variation in the scale of CNV, but the memory and computational costs required for this method are too burdensome.  In addition, Wang~\textit{et al.}~\cite{wang2020automated} proposed a two-stage network based on a deep learning algorithm, which determines the presence or absence of CNV. It is worth noting that, although these methods have achieved impressive results, the imaging differences from multiple device domains and clinical centers continue to challenge the generalizability of existing methods. Therefore, developing a reliable and accurate automatic CNV segmentation algorithm remains a significant challenge.


\subsection{Graph convolutional networks (GCNs)}
In recent years, the exploration of GCNs in computer vision tasks has gained significant attention. By processing graph-structured data, GCNs can effectively capture spatial contextual information for reasoning interaction, achieving state-of-the-art performance in both natural and medical imaging tasks~\cite{wu2020comprehensive,kipf2016semi,liu2020cnn,hu2020class,soberanis2020uncertainty,ma2019attention,xie2021scale}. 
Specifically, graph convolution is typically categorized into two types: feature space graph convolution and coordinate space graph convolution. The former captures interdependencies along the feature map's channels, projecting features into a non-coordinate space~\cite{li2018beyond,chen2019graph,huang2020referring}, whereas coordinate space graph convolution models spatial relationships between pixels to generate consistent predictions across discrete infected regions, projecting features into a new coordinate space~\cite{zhang2019dual,shin2019deep,wu2020ginet,li2017multiple}. For instance, Chen~\textit{et al.}~\cite{chen2019graph} designed a GloRe unit that uses weighted global pooling and broadcasting to map coordinate-interaction space, achieving relational reasoning on small graphs. Zhang~\textit{et al.}~\cite{zhang2019dual} constructed a two-graph convolutional network to model spatial relationships and channel interdependencies, and thereby enhancing information interaction. Li~\textit{et al.}~\cite{li2019pedestrian} employed feature space graph convolution to convert a 2D feature map into a graph structure to capture long-range dependencies. 

In medical imaging, Huang~\textit{et al.}~\cite{huang2021graph} developed a pyramidal global contextual reasoning architecture for multi-scale long-range interactions, while Shin~\textit{et al.}~\cite{shin2019deep} combined GCN and CNN to jointly learn the global structure and local appearance of blood vessel shapes. However, these methods mainly focus on using coordinate space graph convolution to capture spatial relationships between pixels for global information propagation and long-range dependency extraction, often neglecting the task-specific reasoning capabilities of feature space graph convolution, particularly in the context of OCTA-based CNV segmentation tasks.
\renewcommand{\tablename}{TABLE}
\begin{table}[!h]
\centering
\caption{PRESENTATION OF EXPERIMENTAL DATA.}
\resizebox{0.48\textwidth}{!}{
\setlength{\tabcolsep}{0.7mm}
\renewcommand\arraystretch{1.5}
\begin{tabular}{l|ccc|cc} 
\hline
\multicolumn{1}{c||}{} & \multicolumn{1}{c}{$3 \times 3~mm^2$} &\multicolumn{1}{c}{$6 \times 6~mm^2$} & \multicolumn{1}{c||}{$9 \times 9~mm^2$} & \multicolumn{1}{c}{$304 \times 304$} & \multicolumn{1}{c}{$384 \times 384$}\\
\hline 
\multicolumn{1}{c||}{\small{Heidelberg}} & \multicolumn{1}{c}{128} &\multicolumn{1}{c}{$-$} & \multicolumn{1}{c||}{$-$} & \multicolumn{1}{c}{$-$} & \multicolumn{1}{c}{\checkmark}\\
\hline
\multicolumn{1}{c||}{\small{Optovue}} & \multicolumn{1}{c}{$-$} &\multicolumn{1}{c}{40} & \multicolumn{1}{c||}{$-$} & \multicolumn{1}{c}{\checkmark} & \multicolumn{1}{c}{$-$}\\
\hline
\multicolumn{1}{c||}{\small{Zeiss}} & \multicolumn{1}{c}{15} &\multicolumn{1}{c}{17} & \multicolumn{1}{c||}{$-$} & \multicolumn{1}{c}{$-$} & \multicolumn{1}{c}{\checkmark}\\
\hline
\multicolumn{1}{c||}{\small{SVision}} & \multicolumn{1}{c}{$-$} &\multicolumn{1}{c}{11} & \multicolumn{1}{c||}{13} & \multicolumn{1}{c}{$-$} & \multicolumn{1}{c}{\checkmark}\\
\hline\hline
\end{tabular}}
\label{dataset}
\end{table}
\section{Datasets}
\subsection{Datasets acquisition}
The experimental dataset includes four types of OCTA imaging devices. Among them, the Heidelberg, SVision, and Zeiss datasets were collected from Ningbo Eye Hospital in Ningbo, China, while the Optovue dataset was collected from the Eye Center of the Second Affiliated Hospital of Zhejiang University in Hangzhou, China. These datasets included 224 acquisitions centered around the macula, with scan ranges covering $3 \times 3~mm^2$, $6 \times 6~mm^2$, and $9 \times 9~mm^2$, and image resolutions of $304 \times 304$ and $384 \times 384$ pixels. More detailed information of data distribution is provided in Table~\ref{dataset}. The input for the CNN model consisted of en-face images of the avascular complex, spanning from the Outer Plexiform Layer (OPL) to Bruch’s Membrane (BM) and Retinal Pigment Epithelium (RPE), as exemplified in Fig.~\ref{dataset_description}. All image acquisitions were approved by the review board of the Ningbo Eye Hospital and the Eye Center of the Second Affiliated Hospital of Zhejiang University, in accordance with the Helsinki Declaration, and each participant provided written informed consent.
\renewcommand{\tablename}{TABLE}
\begin{table}[!h]
\setlength{\tabcolsep}{2mm}{} 
\renewcommand\arraystretch{1}
\centering
\caption{EXPLANATIONS OF THE NOTATIONS IN THE METHODOLOGY.}
\setlength{\tabcolsep}{0.6mm}
\renewcommand\arraystretch{1.2}
\begin{tabular}{ll} 
\hline
\multicolumn{1}{c}{Symbols}  & \multicolumn{1}{c}{Explanations} \\
\hline 
\multicolumn{1}{l}{\small{$F$}}  & \multicolumn{1}{l}{Feature maps for boundary, shape, and region tasks.}\\
\multicolumn{1}{l}{$F^{'}$} & \multicolumn{1}{l}{Feature maps after $1 \times 1$ convolution.} \\
\multicolumn{1}{l}{\small{$\mathcal{G}$}} & \multicolumn{1}{l}{The graph node embedding as manipulated by $G_\textrm{pro}$.}\\
\multicolumn{1}{l}{\small{T}} & \multicolumn{1}{l}{Cluster center matrix of $K \times C$, with $t_k$ as the $k$-th center.} \\
\multicolumn{1}{l}{\small{$\sigma$}} & \multicolumn{1}{l}{Scale matrix, where $\sigma_k$ corresponds to the $k$-th cluster center.}  \\
\multicolumn{1}{l}{\small{$f_{i}$}} & \multicolumn{1}{l}{Eigenvector in $F$, where $i$ denotes a specific vector.}  \\
\multicolumn{1}{l}{\small{$S_{k}^{j}$}} & \multicolumn{1}{l}{Soft assignment of feature vector $f_j$ to clustering center $t_k$.}  \\
\multicolumn{1}{l}{\small{$A$}} & \multicolumn{1}{l}{The adjacency matrix of the graph.}  \\
\multicolumn{1}{l}{\small{$g_{k}$}} & \multicolumn{1}{l}{Final normalized representation of the $k$-th node.}  \\
\multicolumn{1}{l}{\small{$g^{\ast}_{k}$}} & \multicolumn{1}{l}{Weighted residual of the $k$-th node.}  \\
\multicolumn{1}{l}{\small{$W$}} & \multicolumn{1}{l}{Weight parameter to control inter-graph information flow.}  \\
\multicolumn{1}{l}{\small{$\Phi(\cdot)$}} & \multicolumn{1}{l}{Nonlinear activation function for complex feature learning.}  \\
\multicolumn{1}{l}{\small{$M$}} & \multicolumn{1}{l}{Weight matrix to adjust the weights of the input features.}  \\
\multicolumn{1}{l}{\small{$\mathcal{B}$, $\mathcal{S}$}} & \multicolumn{1}{l}{Boundary and shape maps mapped by fully connected layers.}  \\
\multicolumn{1}{l}{\small{$F_\mathcal{B}$, $F_\mathcal{S}$}} & \multicolumn{1}{l}{Boundary and shape features from channel attention on $F^{\text{reg}}$.}  \\
\multicolumn{1}{l}{\small{$E^B_{m,i}$}} & \multicolumn{1}{l}{Boundary embeddings, encoding node boundary information.}  \\
\multicolumn{1}{l}{\small{$E^S_{m,i}$}} & \multicolumn{1}{l}{Shape embeddings, encoding node shape information.}  \\
\multicolumn{1}{l}{\small{$f_w$}} & \multicolumn{1}{l}{Nonlinear function that captures relationships between nodes.}  \\
\multicolumn{1}{l}{\small{$f_\eta$}} & \multicolumn{1}{l}{Function for learning node embeddings.}  \\ \hline 
\end{tabular}
\label{Natations}
\end{table}
\subsection{Manual annotation protocol}
The imaging conditions and vascular morphology of CNV can vary significantly across subjects, often leading to issues such as interference from projection artifacts and blurred vessel boundaries, which complicate manual annotations of CNV regions and vessels. The data annotation process was conducted as follows: two ophthalmologists performed the initial CNV annotation using ITK-SNAP. To address any ambiguity, 3D OCTA image visualization was referenced to assist both experts in verifying the CNV lesions. After completing the annotations, a cross-validation procedure was carried out by the two experts to ensure consistency. In cases of disagreement, the annotations were reviewed by a senior ophthalmologist for confirmation. Finally, a second senior ophthalmologist conducted a thorough review and made final revisions. 
Notably, the CNV dataset provides both lesion region and vessel labels and is currently the only publicly accessible CNV dataset. The detailed information of the dataset, along with their manual annotations, is available on GitHub\footnotemark[\value{footnote}].
\begin{figure}[!t]
\centering{
\includegraphics[width=1\linewidth]{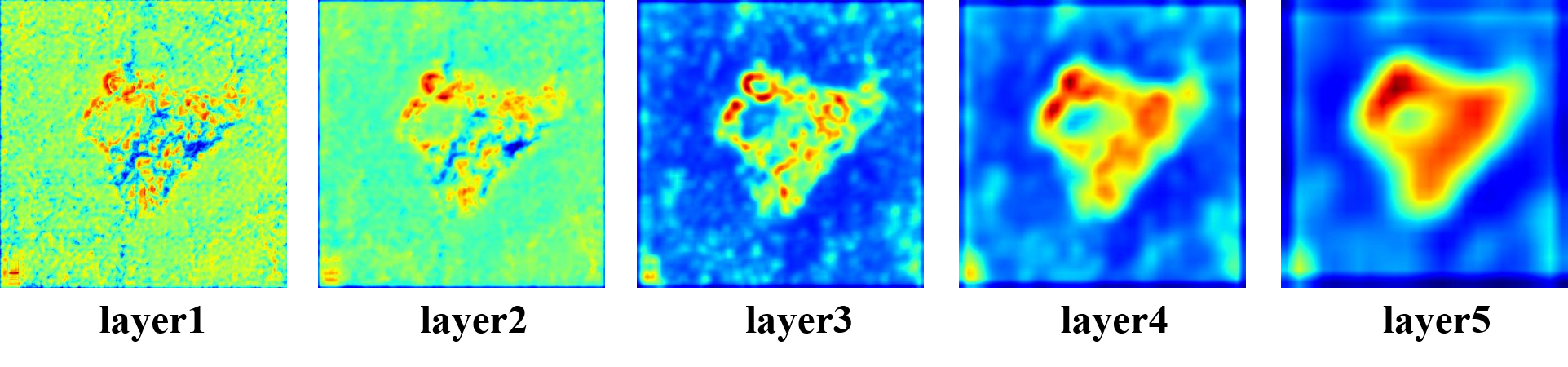}}
\caption{Feature details of each of the 5 encoder layers.}
\label{encoder}
\end{figure}
\begin{figure*}[!t]
\centering
\includegraphics[ width=0.9\linewidth]{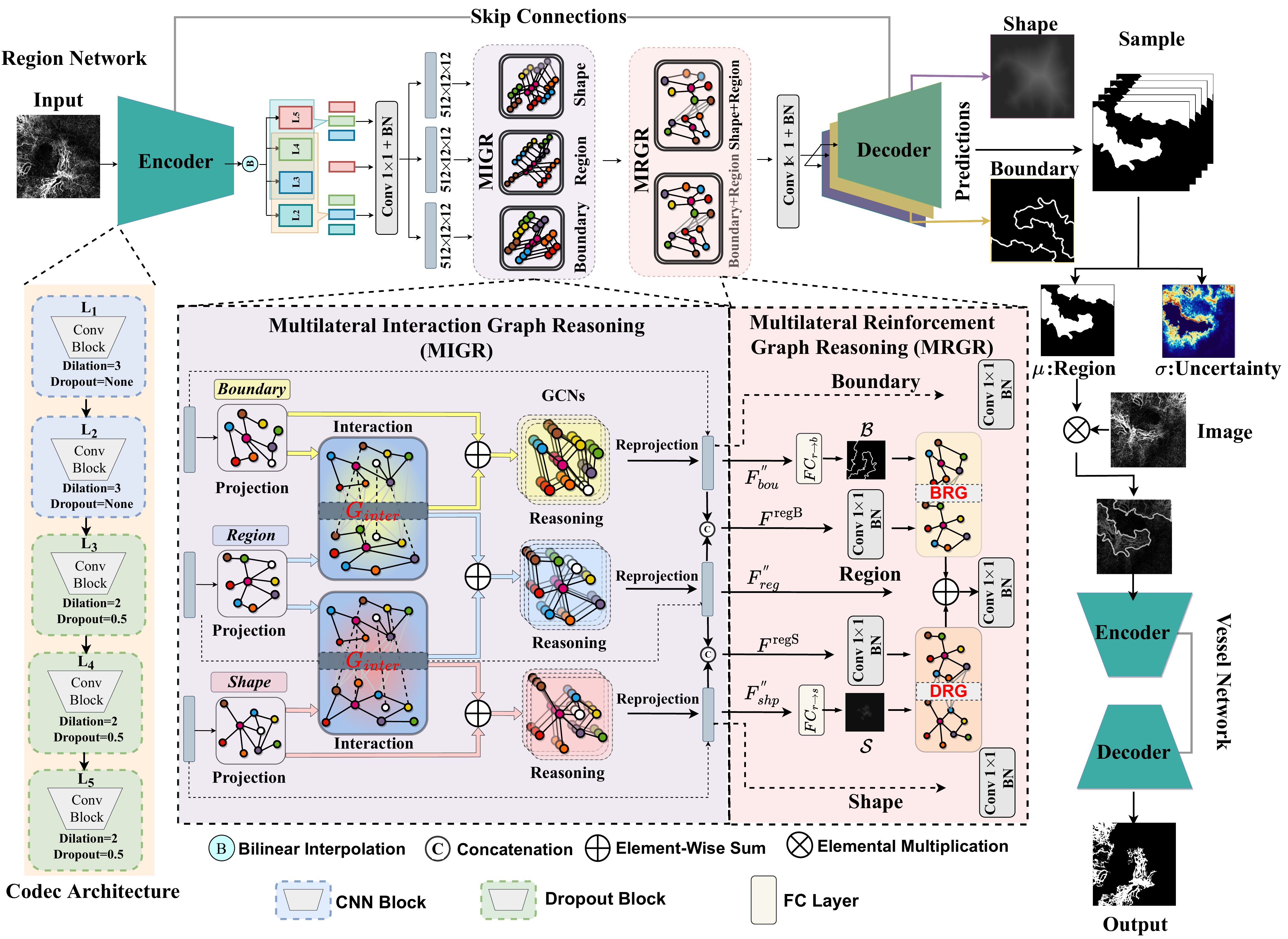}
\caption{Overview of the MTG-Net architecture: It includes a CNN backbone, MIGR for graph interaction, MRGR for graph enhancement, and an uncertainty map for loss. The network is serially structured, with region segmentation feeding into the vessel segmentation branch.}
\label{MTL_image}
\end{figure*}

\section{Methodology}

The proposed MTG-Net consists of three major components: the cascaded CNV region and vessel segmentation networks, Multilateral Interaction Graph Reasoning (MIGR) module and Multilateral Reinforcement Graph Reasoning (MRGR) module, as shown in Fig.~\ref{MTL_image}. We will detail each component in the following sections. In addition, the symbols in the methodology are explained in Table~\ref{Natations}.

\subsection{CNV segmentation network}
In this section, we establish the overall CNV segmentation process, which consists of two main procedures: CNV region segmentation and vessel segmentation. First, the CNV regions are segmented via a multi-task region codec framework. The resulting segmentation maps are then combined with the original input image to serve as input for the subsequent vessel segmentation.
As shown in Fig.~\ref{MTL_image}, the region segmentation framework includes tasks for reconstructing boundary, region, and shape graphs. The subsequent vessel segmentation framework employs the same encoder-decoder structure as the region segmentation procedure.
Specifically, inspired by the two-layer nested U-shape structure of U$^{2}$Net~\cite{qin2020u2}, MTG-Net incorporates a nested U-shape network within each convolutional block. The network consists of five layers, with the internal nested network also starting with five layers. This depth is adjusted by adding or removing one layer at each upsampling or downsampling step. The convolutional kernels within the nested network use dilated convolutions, with a dilation rate of 3 for the first two layers and 2 for the last three layers. This design captures more contextual information at different scales and reduces the computational burden.
To explore the interrelationships among the three tasks, we project the encoded feature vectors into the graph domain for efficient relational reasoning. After reasoning, the features are projected back into the feature space for prediction. 
Considering the hierarchical nature of features extracted by the encoder, the shallow layers typically capture rich boundary details, while the deeper layers focus on representing complex shape structures. As shown in the Fig.~\ref{encoder}. We designate the five layers of the encoder as $L_\textrm{1}$, $L_\textrm{2}$, $L_\textrm{3}$, $L_\textrm{4}$ and $L_\textrm{5}$, as shown in Fig.~\ref{MTL_image}. The boundary mapping is formed by concatenating $L_\textrm{2}$, $L_\textrm{3}$, and $L_\textrm{4}$, while the shape mapping is composed of $L_\textrm{3}$, $L_\textrm{4}$, and $L_\textrm{5}$. $L_\textrm{5}$ serves as the region mapping to ensure that the network can fully learn high-resolution outputs during training. It is worth noting that, due to the inconsistency in feature dimensions extracted from the five stages of the encoder, we employ bilinear interpolation to resize all features to a uniform dimension before performing concatenation. The formal expression is:
\begin{align}
F_\textrm{bou} =&  R \left( L_\textrm{2} \Vert L_\textrm{3} \Vert L_\textrm{4} \right) \\
F_\textrm{shp} =&  R \left( L_\textrm{3} \Vert L_\textrm{4} \Vert L_\textrm{5} \right) \\
F_\textrm{reg} =&  R \left( L_\textrm{5} \right),
\end{align}
where $\Vert$ denotes channel concatenation and $R(\cdot)$ represents the reconstruction function with a $1 \times 1$ convolution followed by batch normalization.

The boundary map is generated using the Canny edge detector~\cite{canny1986computational} on the region mask. Furthermore, the shape map is constructed by applying the signed distance function (SDF) as the shape descriptor~\cite{park2019deepsdf,osher1988fronts,xue2020shape}, allowing to carry rich shape and boundary information. Eventually, we perform a pixel-wise multiplication of the region segmentation results with the input image, in order to focus the segmentation on vascular structures and eliminate interference from non-vascular pixels. Then our approach can effectively mask out the non-vascular areas, improving the segmentation's overall accuracy and reliability.

\subsection{Multilateral interaction graph reasoning (MIGR)}
The MIGR aims to reason about boundary, shape and semantic relationships between region features. The structure, as shown in Fig.~\ref{MTL_image}, involves four key steps which are detailed in the following subsections.

\subsubsection{Graph projection $G_\textrm{pro}$} In this step, the feature maps from the last layer of the encoder are used to construct input features for the region segmentation task. 
Simultaneously, the last four layers of the encoder are concatenated to form input features for the boundary and shape tasks. where $F_\textrm{bou} \in \mathbb{R}^{H \times W \times C}$, $F_\textrm{shp} \in \mathbb{R}^{H \times W \times C}$ and  $F_\textrm{reg} \in \mathbb{R}^{H \times W \times C}$ are reduced to a lower dimension $F^{'}_\textrm{bou} \in \mathbb{R}^{(H \times W) \times C}$, $F^{'}_\textrm{shp} \in \mathbb{R}^{(H \times W) \times C}$ and $F^{'}_\textrm{reg} \in \mathbb{R}^{(H \times W) \times C}$ using a $1 \times 1$ convolution. The reduced feature is then transformed into a graph node embedding $\mathcal{G} \in \mathbb{R}^{C \times K}$ utilizing the projection function $G_\textrm{pro}$. We use a clustering center matrix $T \in \mathbb{R}^{K \times C}$ and a scale matrix $\sigma \in \mathbb{R}^{K \times C}$ to parameterize $G_\textrm{pro}$, where we define each column $t_{k}$ with $k \in \{1,\dots,K\}$ to specify a learnable clustering center for the $k$-th node, and where $K$ represents the number of nodes in $\mathcal{G}$. In our method, $K$ is set to 12. The scale matrix $\sigma$ determines the scale of the clustering centers, $\sigma_{k}$  is the column vector of $\sigma$, with each column $\sigma_{k}$ representing the scale associated with the $k$-th clustering center. 
For each node, the feature vector $f_{i}$ with $i \in (H \times W)$ (representing its low-level feature information) and the learnable clustering center $t_{k}$ (representing its abstract information) are considered.
Consequently, the affinity between these feature vectors and each clustering center is computed using a soft assignment strategy $S_{k}^{j}$, which assigns weights to each node’s contribution and can be expressed as follows:
\begin{equation}
 S_{k}^{i} = \frac{\exp(-\|(f_{i}-t_{k}) \oslash \sigma_{k}\|_{2}^{2}/2)}{\sum_{j}^{K} \exp(-\|(f_{i}-t_{j})\oslash \sigma_{j}\|_{2}^{2}/2)},
\end{equation}
where $\oslash$ represents the element-wise division, $t_{j}$ and $\sigma_{j}$ represent cluster centers other than the $k$-th cluster center and the associated scale, respectively. By calculating the differences between the input features $f_{i}$ and the cluster center $t_{k}$, we then obtain the weighted average residuals ${g}^{\ast}_{k}$ of these differences. This process gradually emphasizes feature vectors with high impact, refining the final node representation ${g}_{k}$, which is formulated as follows:
\begin{equation}
g_{k}=\frac{{g}^{\ast}_{k}}{\|{g}^{\ast}_{k}\|_{2}}\text{~and~}{g}^{\ast}_{k}=\frac{1}{\sum_{i}S_{k}^{i}} \sum_{i}S_{k}^{i}(f_{i}-t_{k})\oslash \sigma_{k}.
\end{equation}
The adjacency matrix $A$ of a graph $ \mathcal{G}$ is obtained by measuring the correlation between node representations, defined as $A  =  \mathcal{G}^{T} \times \mathcal{G}$.
\subsubsection{Graph interaction $G_\textrm{inter}$}
\begin{figure}[t]
\centering{
\includegraphics[width=0.9\linewidth]{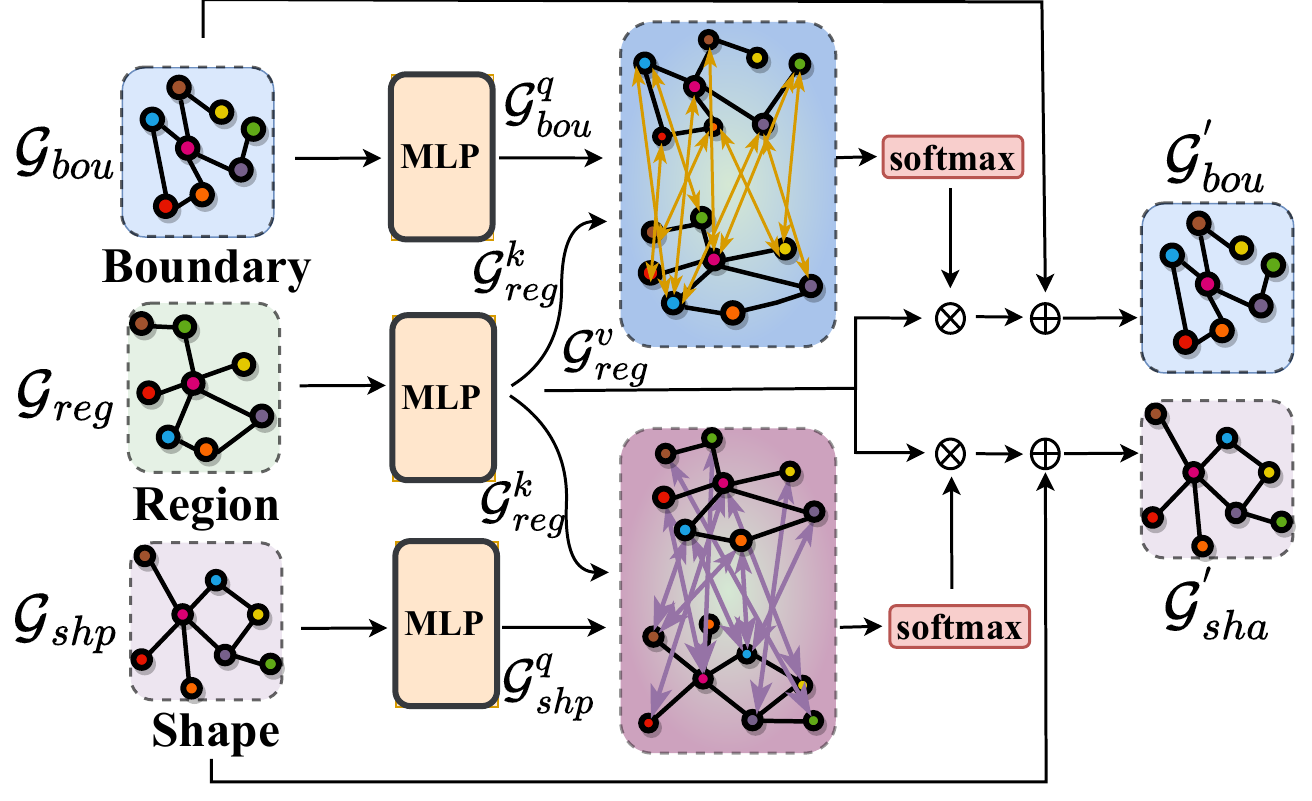}
}
\caption{Schematic diagram of the $G_\textrm{inter}$: After passing through the MLP, input features are passed to boundary-region and shape-region interactions to get outputs separately.}
\label{GII}
\end{figure}

The $G_\textrm{inter}$ models the inter-graph interactions among the boundary, region, and shape graphs ($\mathcal{G}_\textrm{bou}$, $\mathcal{G}_\textrm{reg}$ and $\mathcal{G}_\textrm{shp}$) constructed via $G_\textrm{pro}$, guiding the transfer of intra-graph information from $\mathcal{G}_\textrm{reg}$ to $\mathcal{G}_\textrm{bou}$ and $\mathcal{G}_\textrm{shp}$. Specifically, as shown in Fig.~\ref{GII}, these graphs are first transformed by an MLP, converting graph $\mathcal{G}_\textrm{reg}$ into a key graph $\mathcal{G}_\textrm{regK}$ and a value graph $\mathcal{G}_\textrm{regV}$, and converting graphs $\mathcal{G}_\textrm{bou}$ and $\mathcal{G}_\textrm{shp}$ into query graphs $\mathcal{G}_\textrm{bouQ}$ and $\mathcal{G}_\textrm{shpQ}$, respectively. Subsequently, the adjacency matrix $A_{\text{r}\to\text{b}}$ and $A_{\text{r} \to \text{s}}$, which represent the similarity between nodes in $\mathcal{G}_\textrm{reg}$ and those in $\mathcal{G}_\textrm{bou}$ and $\mathcal{G}_\textrm{shp}$, are then calculated via matrix multiplication, where $\text{r} \to \text{b}$ denotes the adjacency matrix between regions and boundaries, and $\text{r} \to \text{s}$ denotes the adjacency matrix between regions and shapes.
As such, the above operations allow the semantic information to be transferred from $\mathcal{G}_\textrm{reg}$ to $\mathcal{G}_\textrm{bou}$ and $\mathcal{G}_\textrm{shp}$ via the interaction of $G_\textrm{inter}$, resulting in enhanced boundary and shape graphs $\mathcal{G}^{'}_\textrm{bou}$ and $\mathcal{G}^{'}_\textrm{shp}$ as follows:
\begin{align}
A_{\text{r} \to \text{b}}  &=  \mathcal{G}^{T}_\textrm{bouQ} \times \mathcal{G}_\textrm{regK}\\
A_{\text{r} \to \text{s}}  &=  \mathcal{G}^{T}_\textrm{shpQ} \times \mathcal{G}_\textrm{regK}
\end{align}
\begin{equation}
\resizebox{0.85\columnwidth}{!}{$
\begin{aligned}
\mathcal{G}^{'}_\textrm{bou}  &=  G_\textrm{inter}(\mathcal{G}_\textrm{reg} \times \mathcal{G}_\textrm{bou})  =  \mathcal{W}(A_{\text{r} \to \text{b}} \times \mathcal{G}^{T}_\textrm{regV}) \oplus \mathcal{G}_\textrm{bou}
\end{aligned}$}
\end{equation}
\begin{equation}
\resizebox{0.85\columnwidth}{!}{$
\begin{aligned}
\mathcal{G}^{'}_\textrm{shp}  &=  G_\textrm{inter}(\mathcal{G}_\textrm{reg}\times \mathcal{G}_\textrm{shp})  =  \mathcal{W}(A_{\text{r} \to \text{s}}\times \mathcal{G}^{T}_\textrm{regV})\oplus \mathcal{G}_\textrm{shp}
\end{aligned}$}
\end{equation}
where $A_{\text{r} \to \text{b}}$, $A_{\text{r} \to \text{s}} \in \mathbb{R}^{K \times K}$, and where $\mathcal{W}$ is used as a weight parameter to regulate the strength of the information transferred from $\mathcal{G}_\textrm{reg}$ to $\mathcal{G}_\textrm{bou}$, $\mathcal{G}_\textrm{shp}$ in $G_\textrm{inter}$.

\subsubsection{Graph reasoning} After performing inter-graph interaction, the resulting graphs $\mathcal{G}^{}_\textrm{bou}$ and $\mathcal{G}^{'}_\textrm{shp}$ are used as inputs to perform intra-graph reasoning and generate enhanced graph representations. Specifically, the adjacency matrix of the graph is first multiplied with the feature matrix to aggregate information from neighboring nodes for each node. Then, a weight matrix is used to linearly transform the aggregated features, mapping them to a new feature space. Finally, a non-linear activation function is applied, allowing the model to learn more complex feature mappings. Additionally, $\mathcal{G}_\textrm{reg}$ is also input into the graph convolution for reasoning and training. The above procedure can be written as:
\begin{align}
\mathcal{G}^{''}_\textrm{bou} &=\Phi (A_\textrm{bou}\,\mathcal{G}^{'}_\textrm{bou}\,M_\textrm{bou})\\ 
\mathcal{G}^{''}_\textrm{shp} &= \Phi (A_\textrm{shp}\,\mathcal{G}^{'}_\textrm{shp}\,M_\textrm{shp}) 
\end{align}
where $A$ is the adjacency matrix of the corresponding graph, $M$ represents the learnable parameters of the graph convolution layer, and $\Phi(\cdot)$ s the nonlinear activation function.

\subsubsection{Graph reprojection} The final step of the MIGR is graph reprojection. In this phase, the soft assignment strategy used in the graph projection step is re-applied to allocate node features. Specifically, we denote the allocation matrix as $Q_\textrm{bou}$, $Q_\textrm{shp}$ and $Q_\textrm{reg}$. The allocation matrix is multiplied by the augmented graph node feature matrices $\mathcal{G}^{''}_\textrm{bou}$, $\mathcal{G}^{''}_\textrm{shp}$ and $\mathcal{G}^{''}_\textrm{reg}$. Finally, the original features $F^{'}_\textrm{bou}$, $F^{'}_\textrm{shp}$ and $F^{'}_\textrm{reg}$ are added to preserve and integrate the initial feature information into the final output. The graph reprojection can be expressed as:
\begin{align}
F^{''}_\textrm{bou} &= Q_\textrm{bou} \mathcal{G}^{''^T}_\textrm{bou} \oplus F^{'}_\textrm{bou}\\ 
F^{''}_\textrm{shp} &= Q_\textrm{shp} \mathcal{G}^{''^T}_\textrm{shp} \oplus F^{'}_\textrm{shp}\\
F^{''}_\textrm{reg} &= Q_\textrm{reg} \mathcal{G}^{T}_\textrm{reg} \oplus F^{'}_\textrm{reg}
\end{align}
where $Q_\textrm{bou}$ ,$Q_\textrm{shp}$ and $Q_\textrm{reg} \in \mathbb{R}^{(H \times W) \times K}$. $F^{''}_\textrm{bou}$ , $F^{''}_\textrm{shp}$ and $F^{''}_\textrm{reg} \in \mathbb{R}^{(H \times W) \times C}$ are the enhanced feature maps for boundary , shape and region respectively.

\begin{figure}[t]
\centering{
\includegraphics[width=8.5cm]{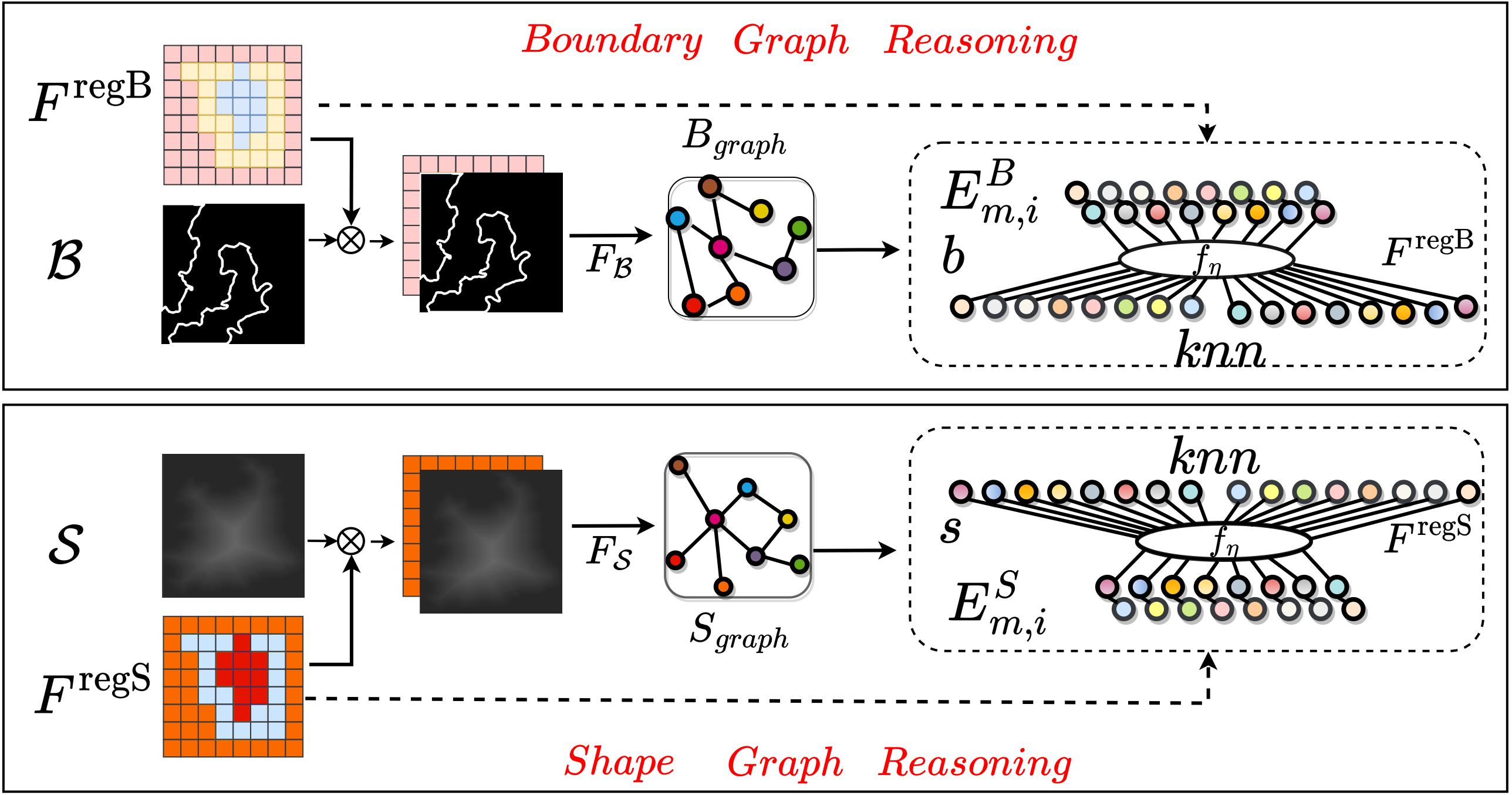}
}
\small
\caption{Schematic diagram of the BRG and SRG: Mining useful information from regions and shapes to guide the final region segmentation.}
\label{BRGDRG}
\end{figure}
\subsection{Multilateral reinforcement graph reasoning (MRGR)}
The primary function of MRGR is to leverage the relationships between boundaries, shapes, and regions to extract valuable information that aids in guiding region segmentation. The structure is shown in the Fig.~\ref{MTL_image}. First, we directly fuse $F^{''}_\textrm{bou}$ and $F^{''}_\textrm{shp}$ with $F^{''}_\textrm{reg}$ to generate the enhanced feature maps $F^\textrm{regB}$ and $F^\textrm{regS}$, respectively. 
These maps are then provided to the BRG and SRG steps for updating.
In addition, $F^{''}_\textrm{bou}$ and $F^{''}_\textrm{shp}$ will also be fed into their respective classifiers to get the segmented boundary map $\mathcal{B}$ and shape map $\mathcal{S}$, which are inputted into the BRG and SRG.

\subsubsection{Boundary and shape node construction} We first apply a fully connected layer to map $F^{''}_\textrm{bou}$ and $F^{''}_\textrm{shp}$ onto their respective boundary maps $\mathcal{B}$ and shape maps $\mathcal{S}$. Subsequently, we model the boundary-related features in $F^\textrm{regB}$ and shape-related features in $F^\textrm{regS}$ by performing a weighted multiplication operation along the channel dimension. This operation essentially functions as an attention mechanism to emphasize important channel features, as illustrated in Fig.~\ref{BRGDRG}. Specifically, the boundary-related features $F_\mathcal{B}$ are obtained by element-wise multiplying $\mathcal{B}$ with $ F^\textrm{regB}$, which can be expressed as $F_\mathcal{B} = \mathcal{B} \odot F^\textrm{regB}$. Similarly, the shape-related features $F_\mathcal{S}$ are derived by element-wise multiplying $\mathcal{S}$ with $F^\textrm{regS}$, represented as $F_\mathcal{S} = \mathcal{S} \odot F^\textrm{regS}$, where $\mathcal{B},\mathcal{S} \in \mathbb{R}^{H \times W \times 1}$, $\odot$ denotes the channel-wise multiplication operation. Finally, the $G_\textrm{pro}$ operation is applied to convert $F_\mathcal{B}$ and $F_\mathcal{S}$ into $k$ node embedding representations based on boundaries and shapes. These embeddings are denoted as $b =\left \{b_\textrm{1},b_\textrm{2}\dots ,b_\textrm{k}\right \}$ for boundary priori and $s=\left \{s_\textrm{1},s_\textrm{2}\dots ,s_\textrm{k}\right \}$ for shape priori. 


\subsubsection{Boundary-enhanced and shape-enhanced Graphs} In this section we construct two graphs, named boundary-enhanced graph (BRG) and shape-enhanced  graph (SRG),  both of which explicitly encode boundary and shape information into nodes and enhance their representations. These k-nearest-neighbor (k-NN) graphs are built based on the similarity of data points in the feature space, connecting $b$ and $s$ with $F^\textrm{regB}$ and $F^\textrm{regS}$. Specifically, for each group of central nodes, boundary-enhanced nodes and shape-enhanced nodes, we first perform element-level subtraction operations to compute the differences between the nodes. Meanwhile, by using the convolution operation, we further introduce nonlinear operations to optimize boundary and shape feature representations. This produces the boundary embedding $E^{B}_{m,i}$ and shape embedding $E^{S}_{m,i}$ as follows
\begin{align}
E^{B}_{m,i} &= f_{w}(f^\textrm{regB}_{i} -b_{m}), \\
E^{S}_{m,i} &= f_{w}(f^\textrm{regS}_{i} -s_{m}), 
\end{align}
where $m \in \{1,\dots,K\}$ and $f_{w}$ denotes the nonlinear function with learnable parameter $w$, used to capture the relationships between the node and its edge-supportive nodes as well as shape-supportive nodes. The feature vectors $f^\textrm{regB}_{i} \in$ $F^\textrm{regB}$ and $f^\textrm{regS}_{i} \in F^\textrm{regS}$ are regarded as the central nodes, and where $b_{m}$ indicates the boundary-supporting node and $s_{m}$ is the shape-supporting node. Thus, for the $i_\textrm{th}$ feature vector, the enhanced output by BRG and SRG are given by:
\begin{align}
\hat{f}^\textrm{regB}_{i} &= \max_{m} f_{\eta}(f^\textrm{regB}_{i},E^{B}_{m,i}), \\
\hat{f}^\textrm{regS}_{i} &= \max_{m}f_{\eta}(f^\textrm{regS}_{i},E^{S}_{m,i}),
\end{align}
where $\hat{f}^\textrm{regB}_{i}$, $\hat{f}^\textrm{regS}_{i}$ $\in$ $\hat{F}^\textrm{regB}$, $\hat{F}^\textrm{regS}$ are the updated representations and $f_{\eta}$ is a node embedding learning function with learnable parameters $\eta$. By integrating the augmented features from BRG and SRG with the initial region features, both boundary and shape information can be embedded into the region segmentation. Through iterative training, the accuracy of region segmentation is ultimately improved.

\subsection{Uncertainty estimated loss}
Although convolutional neural networks (CNNs) are effective in extracting CNV features, boundary blurring due to artifacts and noise often results in blurred lesion boundaries, which makes prediction less reliable. To address this issue, we integrate the probabilistic prediction mechanism of Bayesian neural networks (BNNs), which enhances the model's robustness against such disturbances and improves overall accuracy. In addition, Given the challenges of low computational efficiency and slow convergence in traditional Bayesian methods, we adopt the Monte Carlo Dropout (MC-Dropout) approach proposed by~\cite{gal2015bayesian} to approximate BNNs. In our method, dropout mechanisms with a rate of $0.5$ are applied to the lower three layers of both the encoder and decoder.

Specifically, for a given input image $I$, during forward propagation, a portion of neurons is randomly dropped, and the predicted distribution $\hat{P}_z$ from the $z$-th sampling is defined by:
\begin{align}
\hat{P}_{z} = f_{\theta_{z}}(I), \quad  z = 1,2,\cdots,Z
\end{align}
where $\theta_z$ represents the network parameters obtained through the dropout mechanism, and $z$ is set as 10. During the testing phase, a set of predictions is generated through multiple forward propagations. By statistically analyzing these predictions, we compute the mean $P$ and variance $V$, which serve as the final prediction output and uncertainty estimate, respectively. Here, $V$ denotes a pixel-wise uncertainty map of the same spatial dimension as the input image, where each element $V_i$ corresponds to the uncertainty at the $i$-th pixel location. These are formally expressed as:
\begin{align}
&P=\frac{1}{Z} \sum_{z=1}^{Z} \hat{P}_{z}, \\
&V=\frac{1}{Z} \sum_{z=1}^{Z}\left(\hat{P}_{z}-P \right)^{2} .
\end{align}
However, since this dropout operation is inherently discontinuous, it does not provide effective gradient information for the parameters in the neural network. As a result, it becomes challenging to directly leverage uncertainty estimates for network optimization during training. To address this issue, we designed a differentiable uncertainty estimation loss function $\mathcal{L}_{uce}$ as a surrogate to aid network optimization. This loss function is designed to minimize errors in regions with high uncertainty within the prediction results, ultimately enhancing both accuracy and reliability in CNV segmentation. The $\mathcal{L}_{\rm uce}$  loss is formally expressed as:
\begin{eqnarray}
\resizebox{0.4 \textwidth}{!}{$
\begin{aligned}
\mathcal{L}_\textrm{uce}=-\frac{1}{Z}\sum_{i = 1}^{Z} &\bigg(\big(1+ \textrm{Normalize}(V_i)\big)\cdot \\
&\big(X_i\log(\hat{X}_i) +(1-X_i)\log(1-\hat{X}_i)\big)\bigg),    
\end{aligned}
$}
\end{eqnarray}
$X_{i}$ and $\hat{X}_{i}$ are the ground truth and predicted values for pixel $i$, respectively, and $V_i$ is the uncertainty of $\hat{X}_{i}$.
\begin{figure*}[!h]
    \centering{
    \includegraphics[width=0.95\linewidth]{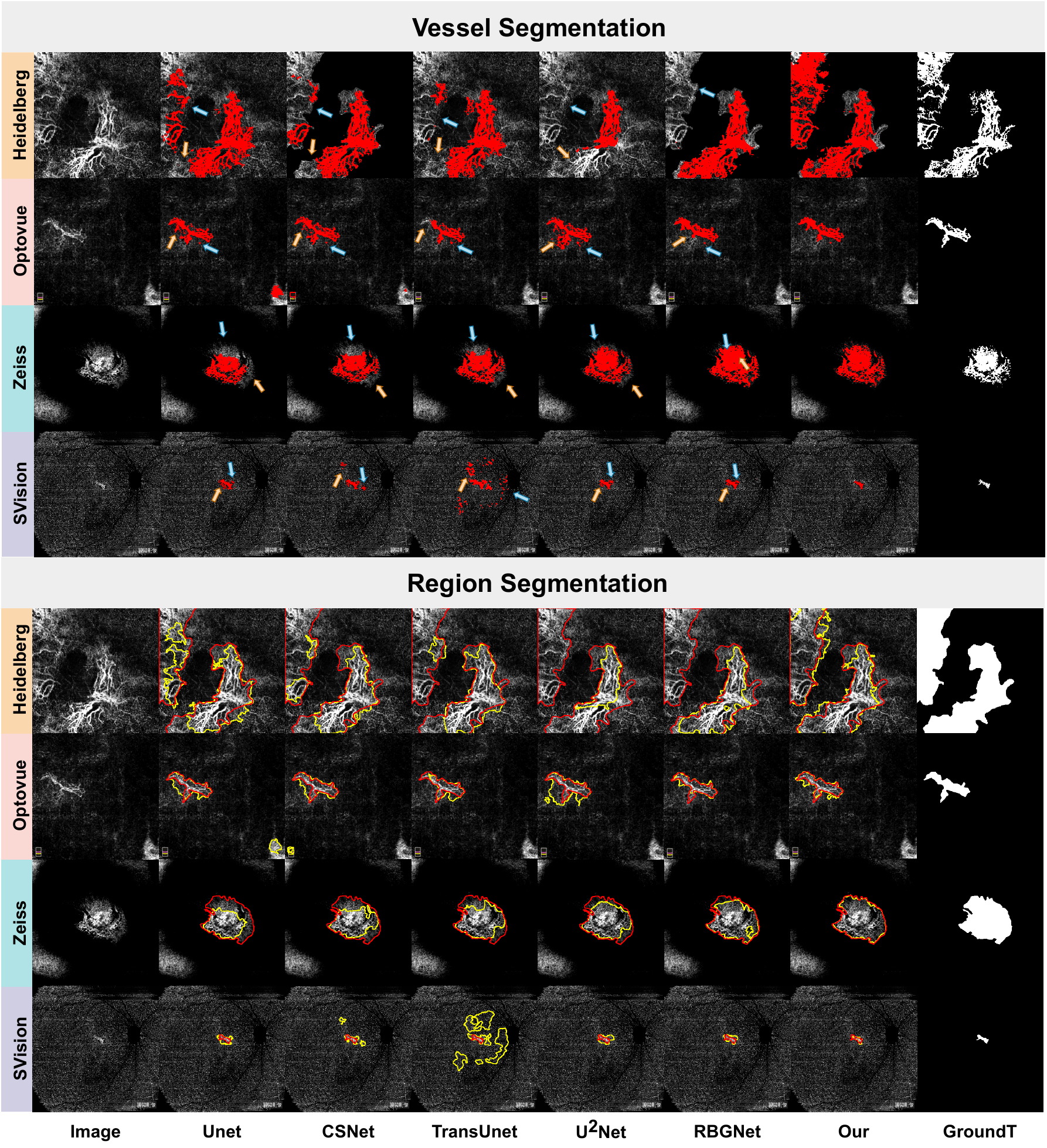}
    }
        \caption{Qualitative comparisons of MTG-Net with other methods are shown. Red indicates the vessel segmentation results from competing methods, while red lines show the region contours from the ground truth. Yellow highlights the segmentation results produced by different methods.}
    \label{fig-compare—image}
\end{figure*}

\subsection{Loss function}
Our approach optimizes four loss components: $\mathcal{L}_{reg}$ for region segmentation, $\mathcal{L}_{shp}$ for shape learning, $\mathcal{L}_{bou}$ for boundary constraint, and $\mathcal{L}_{ves}$ for vessel segmentation. The losses include Binary Cross-Entropy (BCE), Mean Squared Error (MSE), Dice Loss, and a proposed uncertainty-weighted loss, defined as follows:
\begin{align}
\mathcal{L}_{reg} = \mathcal{L}^\textrm{r}_{dice}+ \mathcal{L}^\textrm{r}_{uce}  \\
\mathcal{L}_{bou} =  \mathcal{L}^\textrm{b}_{dice}+ \mathcal{L}^\textrm{b}_{uce} \\
\mathcal{L}_{shp} =  \mathcal{L}^\textrm{s}_{mse}+ \mathcal{L}^\textrm{s}_{uce} \\
\mathcal{L}_{ves} = \mathcal{L}^\textrm{v}_{dice}+ \mathcal{L}^\textrm{v}_{uce}
\end{align}
where $\mathcal{L}^{r}_{uce}$, $\mathcal{L}^{b}_{uce}$, $\mathcal{L}^{s}_{uce}$, and $\mathcal{L}^{v}_{uce}$ represent the uncertainty losses for region, boundary, shape, and vessel tasks, respectively.
To handle weight adjustment in the loss combination for region segmentation, we normalize the uncertainty of each task as adaptive weights. These coefficients are adaptively updated every 50 training iterations, with a total of 6 adjustments. The weight coefficients are defined as:
\begin{align}
\lambda_{i} = \frac{V_{i}}{V_{reg} + V_{bou} + V_{shp}}
\end{align}
where $V_i$ denotes the uncertainty estimation result, and $i \in \{1,2,3\}$ respectively represents the region, boundary and shape tasks $V_{reg}, V_{bou}$ and $V_{shp}$. The total loss is defined as:
\begin{equation}\small
\mathcal{L}_{total}=(\lambda_{reg}\mathcal{L}_{reg}+\lambda_{bou}\mathcal{L}_{bou}+\lambda_{shp}\mathcal{L}_{shp})+\mathcal{L}_{ves},
\end{equation}

\section{Experiments}
\subsection{Experimental settings}
The model training was carried out on a workstation equipped with an NVIDIA GeForce GTX 3090, using the PyTorch framework. Data augmentation included random flipping with angles uniformly sampled from the range of $\left[-45^{\circ}, 45^{\circ}\right]$. The Adam optimizer was applied with a learning rate of 0.0001 and a weight decay set to $\textrm{1e-4}$.
The model was trained over 300 iterations on a mixed dataset comprising images from Heidelberg, Optovue, SVision, and Zeiss devices, with a batch size of 4. To minimize the potential risk of data leakage, the mixed dataset was partitioned into a training set (60\%), validation set (10\%), and test set (30\%). Additionally, to further assess the model’s robustness and generalization, we performed experiments with re-partitioned datasets. Specifically, data from one imaging device was selected as the test set, while data from the remaining three devices was used for training. For example, the dataset containing only Heidelberg device images was used as the test set, and the other three device datasets were used to evaluate the model's performance.
In all experimental settings, the input image resolution was consistently adjusted to $384 \times 384$ pixels. Considering the impact of data volume on model training, uncertainty assessment sampling was performed over 15 iterations for the mixed dataset and 10 iterations for cross-validation with standalone datasets. The evaluation of MTG-Net is divided into two key components. The first assesses the segmentation of the CNV region, while the second focuses on vessel segmentation. To comprehensively measure the segmentation performance in both tasks, Dice, IoU, Precision and Recall are calculated as pixel-level evaluation across different tasks.
\renewcommand{\tablename}{TABLE}
\begin{table*}[!h]

\setlength{\tabcolsep}{0.5mm}{} 
\renewcommand\arraystretch{1}
\caption{COMPARATIVE EXPERIMENTAL RESULTS OF CNV SEGMENTATION IN MIXED DATASETS.}
\centering
\footnotesize
\begin{tabular}{l|cccc|cccc}
\toprule

\multicolumn{1}{c}{\multirow{2}{*}{Methods}} & \multicolumn{4}{c}{Region segmentation} & \multicolumn{4}{c}{Vessel segmentation}\\ 
\cmidrule{2-9}

\multicolumn{1}{c}{~} & DICE  & IoU & Recall   &\multicolumn{1}{c|}{Pre}  &  DICE  & IoU & Recall &\multicolumn{1}{c}{Pre}\\ 
\midrule

U-Net~\cite{ronneberger2015u} & 0.8086$_{(0.1649)}$  & 0.7045$_{(0.1915)}$   & 0.8347$_{(0.2154)}$  & 0.8497$_{(0.1346)}$   & 0.8309$_{(0.1399)}$   & 0.7290$_{(0.1572)}$  & 0.8960$_{(0.1687)}$   &0.8166$_{(0.1308)}$ \\
CE-Net~\cite{gu2019net} & 0.8473$_{(0.1231)}$   & 0.7522$_{(0.1622)}$   & 0.8870$_{(0.1407)}$ & 0.8405$_{(0.1506)}$  & 0.8262$_{(0.1166)}$   & 0.7184$_{(0.1483)}$    & 0.9172$_{(0.1024)}$ & 0.7746$_{(0.1487)}$  \\
CS-Net~\cite{mou2019cs} & 0.8194$_{(0.1526)}$   & 0.7184$_{(0.1905)}$   & 0.8551$_{(0.1774)}$ & 0.8390$_{(0.1731)}$ & 0.8418$_{(0.1220)}$   & 0.7424$_{(0.1504)}$    & 0.9081$_{(0.1204)}$   & 0.8205$_{(0.1633)}$ \\
TransUNet~\cite{chen2021transunet} & 0.8126$_{(0.1464)}$   & 0.7047$_{(0.1783)}$   & 0.8672$_{(0.1708)}$ & 0.8070$_{(0.1687)}$ & 0.8291$_{(0.1217)}$   & 0.7246$_{(0.1599)}$    & 0.9015$_{(0.1249)}$   & 0.8017$_{(0.1671)}$\\ 
SwinUnet~\cite{cao2022swin} & 0.7406$_{(0.1837)}$	&0.6182$_{(0.2108)}$	 &0.7365$_{(0.2189)}$ &0.8374$_{(0.2087)}$ & 0.7059$_{(0.1521)}$   &0.5644$_{(0.1619)}$	 &0.8543$_{(0.1811)}$ &0.6581$_{(0.1830)}$\\ 
U$^{2}$Net~\cite{qin2020u2} & 0.8432$_{(0.1453)}$  & 0.7518$_{(0.1838)}$  & 0.8768$_{(0.1661)}$   &0.8525$_{(0.1593)}$  & 0.8616$_{(0.1389)}$   & 0.7812$_{(0.1661)}$  & 0.9101$_{(0.1257)}$     &0.8497$_{(0.1457)}$\\ 
H2Formert~\cite{he2023h2former} & 0.7503$_{(0.1856)}$   &0.6299$_{(0.2101)}$   & 0.7249$_{(0.2149)}$ &0.8648$_{(0.1636)}$  & 0.7223$_{(0.1511)}$  & 0.5837$_{(0.1701)}$   & 0.8521$_{(0.2037)}$	 &0.6851$_{(0.1747)}$\\ 
MedFormer~\cite{wang2024medformer} & 0.7491$_{(0.1501)}$	&0.6211$_{(0.1763)}$	 &0.7705$_{(0.1722)}$ &0.7979$_{(0.2036)}$ &0.6992$_{(0.1403)}$ &0.5536$_{(0.1473)}$ &0.8863$_{(0.1483)}$  &0.6150$_{(0.1692)}$\\ 
Meng-Net~\cite{meng2021graph} & 0.8164$_{(0.1263)}$  & 0.7201$_{(0.1653)}$  & 0.8679$_{(0.1593)}$  &0.8011$_{(0.1879)}$ & 0.8154$_{(0.1253)}$   & 0.7201$_{(0.1313)}$    & 0.8793$_{(0.1223)}$   &0.7911$_{(0.1707)}$ \\ 
nnUNet~\cite{isensee2021nnu} & 0.8324$_{(0.1669)}$	 &0.7290$_{(0.1983)}$	&0.8747$_{(0.1521)}$&0.8456$_{(0.1599)}$ & 0.8786$_{(0.1768)}$	&0.7522$_{(0.2012)}$	&0.8942$_{(0.1167)}$&0.8474$_{(0.2308)}$ \\ 
D-Persona~\cite{wu2024diversified} & 0.8506$_{(0.1104)}$  & 0.7528$_{(0.1461)}$   & 0.8451$_{(0.1571)}$  & 0.8395$_{(0.1443)}$   & 0.8750$_{(0.1064)}$   & 0.7805$_{(0.1480)}$  & 0.9006$_{(0.1128)}$  & 0.8616$_{(0.1112)}$ \\
MAVNet~\cite{yu2024multi} & 0.8240$_{(0.1754)}$   & 0.7305$_{(0.2044)}$   & 0.8035$_{(0.2087)}$  & 0.8047$_{(0.1765)}$   & 0.7737$_{(0.1656)}$    & 0.6564$_{(0.1920)}$  & 0.7219$_{(0.2190)}$ &0.8659$_{(0.1665)}$  \\ \hline
MF-Net~\cite{meng2021mf} & 0.8213$_{(0.1736)}$  & 0.7260$_{(0.2025)}$   & 0.8835$_{(0.1493)}$  &  0.8180$_{(0.1995)}$  & 0.8170$_{(0.1422)}$   & 0.7104$_{(0.1667)}$  & 0.8861$_{(0.1169)}$  & 0.7958$_{(0.1815)}$ \\
CNV-Net~\cite{vali2023cnv} & 0.8134$_{(0.1769)}$  & 0.7120$_{(0.1948)}$  & 0.8347$_{(0.2211)}$ & 0.8502$_{(0.1739)}$ & 0.8435$_{(0.1572)}$   & 0.7321$_{(0.1634)}$    & 0.8834$_{(0.1947)}$   & 0.8157$_{(0.1772)}$\\ 
RBGNet~\cite{chen2023rbgnet} & 0.8532$_{(0.1223)}$  & 0.7580$_{(0.1604)}$  & 0.8540$_{(0.1484)}$ &0.8511$_{(0.1465)}$ & 0.8371$_{(0.1053)}$   & 0.7310$_{(0.1463)}$    & \textbf{0.9304$_{(0.1314)}$}   &0.7825$_{(0.1227)}$ \\ \hline
Proposed   & \textbf{0.8721$_{(0.0880)}$}  & \textbf{0.7740$_{(0.1291)}$}   & \textbf{0.9068$_{(0.1050)}$} & \textbf{0.8660$_{(0.1401)}$} & \textbf{0.8812$_{(0.0861)}$}  & \textbf{0.7974$_{(0.1254)}$}  & 0.9138$_{(0.0933)}$  &  \textbf{0.8731$_{(0.1348)}$}\\ 
\bottomrule
\end{tabular}
\label{tab-comparison_mix}
\end{table*}

\renewcommand{\tablename}{TABLE}
\begin{table*}[!h]
\setlength{\tabcolsep}{1mm}{} 
\renewcommand\arraystretch{0.8}
\caption{SINGLE-DEVICE REGION SEGMENTATION COMPARISON EXPERIMENT RESULTS AS TEST SET (HEIDELBERG, SVISION).}
\centering
\footnotesize
\begin{tabular}{l|ccc|ccc}
\toprule
\multicolumn{1}{c}{\multirow{3}{*}{Methods}} & \multicolumn{6}{c}{Region segmentation}\\ 
\cmidrule{2-7}

\multicolumn{1}{c}{~} &  \multicolumn{3}{c|}{Heidelberg} &\multicolumn{3}{c}{SVision} \\ 
\cmidrule{2-7}

\multicolumn{1}{c}{~} & Dice & IoU & \multicolumn{1}{c|}{Recall} & Dice & IoU &\multicolumn{1}{c}{Recall} \\ 
\midrule

U-Net~\cite{ronneberger2015u} & 0.6970$_{(0.2163)}$  & 0.6231$_{(0.2298)}$   & 0.7018$_{(0.1961)}$  & 0.6415$_{(0.2379)}$ & 0.5216$_{(0.2631)}$  & 0.7952$_{(0.1987)}$ \\

CE-Net~\cite{gu2019net} & 0.7506$_{(0.1857)}$   & 0.6439$_{(0.2171)}$   & 0.8580$_{(0.1996)}$ & 0.7432$_{(0.1752)}$	&0.6162$_{(0.2031)}$	&0.8699$_{(0.1876)}$  \\

CS-Net~\cite{mou2019cs} & 0.7385$_{(0.2032)}$   & 0.6225$_{(0.2330)}$   & 0.8479$_{(0.1972)}$ & 0.6943$_{(0.1899)}$	&0.5648$_{(0.2094)}$	&0.7897$_{(0.1530)}$ \\

TransUNet~\cite{chen2021transunet} & 0.7406$_{(0.2070)}$   & 0.6256$_{(0.2317)}$   & 0.7631$_{(0.2490)}$ &  0.7572$_{(0.1694)}$	&0.6435$_{(0.1992)}$	&0.8253$_{(0.1567)}$ \\

SwinUnet~\cite{cao2022swin} & 0.7451$_{(0.1501)}$	& 0.6224$_{(0.1763)}$  &0.8357$_{(0.1571)}$ & 0.7461$_{(0.1707)}$	&0.6224$_{(0.2002)}$	&0.8663$_{(0.1434)}$ \\ 

U$^{2}$Net~\cite{qin2020u2} & 0.7555$_{(0.2008)}$  & 0.6445$_{(0.2344)}$  & 0.8199$_{(0.1952)}$ &  0.7717$_{(0.1858)}$   &0.6552$_{(0.2124)}$	 &0.8672$_{(0.1349)}$ \\ 

H2Formert~\cite{he2023h2former} & 0.7334$_{(0.1925)}$	&0.6128$_{(0.2241)}$	 &0.7721$_{(0.1743)}$ & 0.7701$_{(0.1707)}$	&0.6582$_{(0.1971)}$	 &0.7797$_{(0.1609)}$ \\ 

MedFormer~\cite{wang2024medformer} & 0.7160$_{(0.1670)}$	&0.6026$_{(0.1932)}$	 &0.7453$_{(0.2103)}$ &0.7762$_{(0.1622)}$	 &0.6640$_{(0.1892)}$ &0.7703$_{(0.1516)}$ \\ 

Meng-Net~\cite{meng2021graph} & 0.7452$_{(0.1489)}$  & 0.6311$_{(0.1857)}$  & 0.7901$_{(0.1904)}$ &0.7451$_{(0.1757)}$	&0.6224$_{(0.2033)}$	 &0.8357$_{(0.1664)}$ \\ 

nnUNet~\cite{isensee2021nnu} & 0.7625$_{(0.1621)}$	&0.6650$_{(0.1308)}$	&0.8612$_{(0.1945)}$ & 0.7428$_{(0.1432)}$	&0.6278$_{(0.1654)}$	&0.8565$_{(0.1624)}$  \\ 
D-Persona~\cite{wu2024diversified} & 0.7862$_{(0.1677)}$	&0.6758$_{(0.2050)}$	&0.7743$_{(0.2021)}$
&0.7282$_{(0.2462)}$	&0.6213$_{(0.2553)}$	&0.7760$_{(0.2992)}$
\\ 
MAVNet~\cite{yu2024multi}  & 0.7928$_{(0.2032)}$	&0.6942$_{(0.2259)}$	&0.7784$_{(0.2139)}$& 0.7565$_{(0.2348)}$	&0.6549$_{(0.2535)}$	&0.7177$_{(0.2797)}$ \\  \hline

MF-Net~\cite{meng2021mf} & 0.7468$_{(0.1960)}$  & 0.6304$_{(0.2233)}$   & 0.7729$_{(0.2214)}$  & 0.7520$_{(0.1917)}$	&0.6382$_{(0.2116)}$	&0.8012$_{(0.1534)}$\\

CNV-Net~\cite{vali2023cnv} & 0.7311$_{(0.1794)}$  & 0.6334$_{(0.1965)}$  & 0.7898$_{(0.2018)}$ & 0.7343$_{(0.1871)}$	&0.6112$_{(0.2048)}$	&0.7924$_{(0.1534)}$ \\ 

RBGNet~\cite{chen2023rbgnet} & 0.7885$_{(0.1550)}$  & 0.6820$_{(0.1863)}$   & 0.7730$_{(0.1889)}$  & 0.7820$_{(0.1478)}$	&0.6696$_{(0.1733)}$	 &0.8621$_{(0.1477)}$  \\  \hline
Proposed   & \textbf{0.8143$_{(0.1334)}$}  & \textbf{0.7104$_{(0.1742)}$}   & \textbf{0.8713$_{(0.1102)}$} & \textbf{0.7936$_{(0.1332)}$}	 & \textbf{0.6769$_{(0.1674)}$} & \textbf{0.8713$_{(0.1483)}$} \\ \bottomrule
\end{tabular}
\label{tab-comparison_test_region_Heidelberg_Svision}
\end{table*}

\renewcommand{\tablename}{TABLE}
\begin{table*}[!h]
\setlength{\tabcolsep}{1mm}{} 
\renewcommand\arraystretch{0.8}
\caption{SINGLE-DEVICE REGION SEGMENTATION COMPARISON EXPERIMENT RESULTS AS TEST SET (ZEISS, OPTOVUE).}
\centering
\footnotesize
\begin{tabular}{l|ccc|ccc}
\toprule
\multicolumn{1}{c}{\multirow{3}{*}{Methods}} & \multicolumn{6}{c}{Region segmentation}\\ 
\cmidrule{2-7}

\multicolumn{1}{c}{~}  &\multicolumn{3}{c|}{Zeiss}  &\multicolumn{3}{c}{Optovue} \\ 
\cmidrule{2-7}

\multicolumn{1}{c}{~} & Dice & IoU & \multicolumn{1}{c|}{Recall} & Dice & IoU &\multicolumn{1}{c}{Recall}   \\ 
\midrule

U-Net~\cite{ronneberger2015u}    & 0.5614$_{(0.2930)}$	& 0.4450$_{(0.2738)}$  & 0.5781$_{(0.2942)}$ & 0.8419$_{(0.1331)}$ & 0.7383$_{(0.1657)}$ & 0.8878$_{(0.1659)}$ \\

CE-Net~\cite{gu2019net}  & 0.7599$_{(0.1910)}$ & 0.6463$_{(0.2211)}$	&0.7218$_{(0.1977)}$ & 0.8866$_{(0.1254)}$ &0.8029$_{(0.1355)}$  &0.8706$_{(0.1451)}$ \\

CS-Net~\cite{mou2019cs}  & 0.7003$_{(0.2249)}$	&0.5798$_{(0.2403)}$	&0.6525$_{(0.2189)}$ & 0.8836$_{(0.1067)}$& 0.7988$_{(0.1503)}$	& 0.8760$_{(0.1642)}$\\

TransUNet~\cite{chen2021transunet}  &0.7811$_{(0.1949)}$	&0.6736$_{(0.2101)}$	&0.7805$_{(0.1930)}$ & 0.8644$_{(0.1024)}$ &0.7744$_{(0.1404)}$	&0.8285$_{(0.1611)}$ \\

SwinUnet~\cite{cao2022swin}  &0.7649$_{(0.1885)}$	&0.6554$_{(0.2005)}$	&0.7843$_{(0.1962)}$
&0.8579$_{(0.1355)}$	&0.7656$_{(0.1309)}$	&0.8323$_{(0.1153)}$\\ 

U$^{2}$Net~\cite{qin2020u2}  &0.7962$_{(0.1818)}$	 &0.6914$_{(0.1971)}$ 	&0.7535$_{(0.2067)}$ & 0.8862$_{(0.0738)}$  &0.8118$_{(0.1114)}$	&0.8741$_{(0.1138)}$ \\ 

H2Formert~\cite{he2023h2former} &  0.7363$_{(0.1712)}$	 &0.6062$_{(0.1789)}$  &0.6301$_{(0.1903)}$ & 0.8551$_{(0.0815)}$	&0.7605$_{(0.1188)}$	&0.8064$_{(0.1227)}$\\ 

MedFormer~\cite{wang2024medformer} & 0.7245$_{(0.1905)}$ &	0.5974$_{(0.2008)}$	&0.6344$_{(0.2051)}$ & 0.8564$_{(0.1083)}$	&0.7622$_{(0.1523)}$	 &0.8138$_{(0.0916)}$\\ 

Meng-Net~\cite{meng2021graph} &  0.7551$_{(0.1824)}$	&0.6362$_{(0.1985)}$	&0.7176$_{(0.2050)}$ & 0.8477$_{(0.1072)}$	&0.7537$_{(0.1506)}$	&0.7774$_{(0.1710)}$\\ 

nnUNet~\cite{isensee2021nnu}  & 0.7916$_{(0.1827)}$	&07012$_{(0.2032)}$	&0.7721$_{(0.2005)}$& 0.8897$_{(0.0828)}$ &0.8167$_{(0.1248)}$  &0.8536$_{(0.1253)}$ \\ 

D-Persona~\cite{wu2024diversified} & 0.8088$_{(0.1723)}$	&0.7030$_{(0.1885)}$	&0.8449$_{(0.1263)}$ &0.8919$_{(0.0991)}$	&0.8280$_{(0.1083)}$	&0.8739$_{(0.1220)}$

\\ 
MAVNet~\cite{yu2024multi}  & 0.7956$_{(0.1677)}$	&0.6862$_{(0.1878)}$	&0.7544$_{(0.1644)}$& 0.8977$_{(0.0867)}$	&0.8211$_{(0.1278)}$	&0.8883$_{(0.1154)}$ \\  \hline
MF-Net~\cite{meng2021mf}  &0.7675$_{(0.1867)}$	&0.6574$_{(0.1993)}$	 &0.6993$_{(0.1948)}$ & 0.8732$_{(0.0945)}$	&0.7872$_{(0.1396)}$	&0.8451$_{(0.1301)}$ \\

CNV-Net~\cite{vali2023cnv} & 0.7487$_{(0.2223)}$	&0.6363$_{(0.2250)}$	&0.6998$_{(0.2178)}$
&0.8476$_{(0.1136)}$	&0.7537$_{(0.1534)}$	&0.7774$_{(0.1699)}$

\\ 
RBGNet~\cite{chen2023rbgnet}  & 0.8049$_{(0.1827)}$	&0.7069$_{(0.2032)}$	&0.7801$_{(0.2005)}$& 0.8986$_{(0.0688)}$	&0.8271$_{(0.1076)}$	&0.8929$_{(0.1037)}$ \\  \hline
Proposed   & \textbf{0.8186$_{(0.1720)}$}  & \textbf{0.7208$_{(0.1928)}$}  & \textbf{0.7827$_{(0.1709)}$} & \textbf{0.9038$_{(0.0635)}$}  & \textbf{0.8304$_{(0.0995)}$}  & \textbf{0.9160$_{(0.1051)}$}\\ \bottomrule
\end{tabular}
\label{tab-comparison_test_region_Zeiss_Optovue}
\end{table*}

\renewcommand{\tablename}{TABLE}
\begin{table*}[t]
\setlength{\tabcolsep}{1.5mm}{} 
\renewcommand\arraystretch{1}
\caption{SINGLE-DEVICE VESSEL SEGMENTATION COMPARISON EXPERIMENT RESULTS AS TEST SET (HEIDELBERG, SVISION).}
\centering
\footnotesize
\begin{tabular}{l|ccc|ccc}
\toprule
\multicolumn{1}{c}{\multirow{3}{*}{Methods}} & \multicolumn{6}{c}{Vessel segmentation}\\ 
\cmidrule{2-7}

\multicolumn{1}{c}{~} &  \multicolumn{3}{c|}{Heidelberg} &\multicolumn{3}{c}{SVision} \\ 
\cmidrule{2-7}

\multicolumn{1}{c}{~} & Dice & IoU & \multicolumn{1}{c|}{Recall} & Dice & IoU &\multicolumn{1}{c}{Recall} \\ 
\midrule

U-Net~\cite{ronneberger2015u}  & 0.7189$_{(0.1912)}$   & 0.6788$_{(0.1789)}$  & 0.7211$_{(0.1647)}$ & 0.6530$_{(0.1976)}$	&0.5351$_{(0.2502)}$	 &0.6846$_{(0.1545)}$   \\
CE-Net~\cite{gu2019net} & 0.7127$_{(0.1789)}$   & 0.5807$_{(0.1967)}$    & 0.8446$_{(0.1995)}$  & 0.7530$_{(0.1752)}$	&0.6259$_{(0.2031)}$	&0.7992$_{(0.1706)}$ \\

CS-Net~\cite{mou2019cs}  & 0.7063$_{(0.2159)}$   & 0.5855$_{(0.2395)}$    & 0.8488$_{(0.2187)}$  &0.7161$_{(0.1599)}$	&0.5918$_{(0.1799)}$	&0.7707$_{(0.1388)}$ \\

TransUNet~\cite{chen2021transunet} & 0.7273$_{(0.2063)}$   & 0.6079$_{(0.2280)}$    & 0.7754$_{(0.2515)}$ &0.7724$_{(0.1342)}$	&0.6634$_{(0.1642)}$	&0.7683$_{(0.1798)}$ \\

SwinUnet~\cite{cao2022swin} &0.7466$_{(0.1403)}$	&0.6201$_{(0.1473)}$	&0.7957$_{(0.1283)}$ &0.7829$_{(0.1537)}$ &	0.6717$_{(0.1889)}$	&0.8156$_{(0.1831)}$  \\ 

U$^{2}$Net~\cite{qin2020u2}  & 0.7542$_{(0.1984)}$   & 0.6421$_{(0.2336)}$    & 0.8029$_{(0.1856)}$  &0.8197$_{(0.1681)}$   &0.7173$_{(0.1984)}$	&0.8596$_{(0.1395)}$ \\ 

H2Former~\cite{he2023h2former} & 0.7524$_{(0.1847)}$	&0.5297$_{(0.1883)}$	 &0.8115$_{(0.1244)}$ & 0.7017$_{(0.1794)}$	&0.5614$_{(0.1757)}$	&0.8575$_{(0.1438)}$ \\ 

MedFormer~\cite{wang2024medformer} & 0.7192$_{(0.1327)}$	&0.6342$_{(0.1498)}$	&0.8012$_{(0.1764)}$ & 0.7309$_{(0.2831)}$	&0.5905$_{(0.2914)}$	 &0.8664$_{(0.1421)}$ \\ 

Meng-Net~\cite{meng2021graph} & 0.7452$_{(0.1223)}$  & 0.6311$_{(0.1857)}$  & 0.7901$_{(0.1904)}$  &0.7466$_{(0.1695)}$	&0.6201$_{(0.1665)}$	 &0.7957$_{(0.1743)}$  \\ 

nnUNet~\cite{isensee2021nnu} &0.7286$_{(0.1340)}$	&0.6222$_{(0.1422)}$	&0.7962$_{(0.2108)}$ & 0.7830$_{(0.1363)}$	&0.6681$_{(0.1663)}$	&0.8953$_{(0.1523)}$ \\ 
D-Persona~\cite{wu2024diversified} & 0.7656$_{(0.1478)}$	&0.6410$_{(0.1745)}$	&0.8557$_{(0.1996)}$ &0.7491$_{(0.2248)}$	&0.6399$_{(0.2322)}$	&0.7747$_{(0.2813)}$\\ 
        
MAVNet~\cite{yu2024multi} &  0.7423$_{(0.2018)}$	&0.6244$_{(0.2166)}$	&0.7544$_{(0.1644)}$& 0.5598$_{(0.2526)}$	&0.4306$_{(0.2419)}$	&0.4446$_{(0.2490)}$ \\  

MF-Net~\cite{meng2021mf} & 0.7070$_{(0.1846)}$   & 0.5750$_{(0.2000)}$  & 0.7930$_{(0.2167)}$ & 0.7368$_{(0.1653)}$	&0.6115$_{(0.1854)}$	&0.7413$_{(0.1496)}$ \\

CNV-Net~\cite{vali2023cnv} & 0.7311$_{(0.1832)}$  & 0.6334$_{(0.2071)}$  & 0.7898$_{(0.2187)}$  & 0.7161$_{(0.1517)}$	&0.5852$_{(0.1817)}$	&0.7480$_{(0.1822)}$ \\ 

RBGNet~\cite{chen2023rbgnet} & 0.7674$_{(0.1442)}$   & 0.6546$_{(0.1661)}$  & 0.7696$_{(0.1895)}$ & 0.8063$_{(0.1387)}$	&0.6979$_{(0.1547)}$	&0.8450$_{(0.1130)}$  \\  \hline

Proposed   & \textbf{0.8109$_{(0.1065)}$}  & \textbf{0.7038$_{(0.1393)}$}  & \textbf{0.8855$_{(0.0872)}$} & \textbf{0.8413$_{(0.1093)}$}	 & \textbf{0.7401$_{(0.1455)}$} & \textbf{0.8833$_{(0.1333)}$}  \\ \bottomrule
\end{tabular}
\label{tab-comparison_test_vessels_Heidelberg_Svision}
\end{table*}

\renewcommand{\tablename}{TABLE}
\begin{table*}[t]
\setlength{\tabcolsep}{1.5mm}{} 
\renewcommand\arraystretch{1}
\caption{SINGLE-DEVICE VESSEL SEGMENTATION COMPARISON EXPERIMENT RESULTS AS TEST SET (ZEISS, OPTOVUE).}
\centering
\footnotesize
\begin{tabular}{l|ccc|ccc}
\toprule
\multicolumn{1}{c}{\multirow{3}{*}{Methods}} & \multicolumn{6}{c}{Vessel segmentation}\\ 
\cmidrule{2-7}

\multicolumn{1}{c}{~} &\multicolumn{3}{c|}{Zeiss}  &\multicolumn{3}{c}{Optovue} \\ 
\cmidrule{2-7}

\multicolumn{1}{c}{~} & Dice & IoU & \multicolumn{1}{c|}{Recall} & Dice & IoU &\multicolumn{1}{c}{Recall}  \\ 
\midrule

U-Net~\cite{ronneberger2015u}   &0.6471$_{(0.3102)}$  	&0.5460$_{(0.2992)}$	 &0.7066$_{(0.3037)}$ &0.8594$_{(0.1102)}$	&0.7638$_{(0.1378)}$	&0.8276$_{(0.1435)}$ \\
CE-Net~\cite{gu2019net} &0.7613$_{(0.1657)}$	&0.6386$_{(0.1819)}$	&0.8573$_{(0.1627)}$ & 0.8770$_{(0.1142)}$	&0.7874$_{(0.1273)}$	&0.8195$_{(0.1363)}$\\

CS-Net~\cite{mou2019cs}  & 0.7991$_{(0.2249)}$	&0.6993$_{(0.2403)}$	&0.8281$_{(0.2189)}$ & 0.8868$_{(0.1086)}$	&0.8024$_{(0.1363)}$	&0.8485$_{(0.1457)}$\\

TransUNet~\cite{chen2021transunet} &0.8158$_{(0.1900)}$	&0.7196$_{(0.1945)}$	&0.8731$_{(0.1848)}$ & 0.8617$_{(0.1389)}$	&0.7675$_{(0.1373)}$	&0.8076$_{(0.1499)}$\\

SwinUnet~\cite{cao2022swin} &0.7876$_{(0.1913)}$	&0.6826$_{(0.1981)}$	& 0.8919$_{(0.1896)}$  &0.8514$_{(0.1150)}$	&0.7534$_{(0.1135)}$	&0.7961$_{(0.1418)}$\\ 

U$^{2}$Net~\cite{qin2020u2}  &  0.8422$_{(0.1754)}$	&0.7545$_{(0.1859)}$	&0.8869$_{(0.1990)}$ & 0.8865$_{(0.0654)}$	&0.8072$_{(0.0996)}$	&0.8592$_{(0.0959)}$\\ 

H2Former~\cite{he2023h2former} &0.7088$_{(0.1454)}$	 &0.5635$_{(0.1322)}$	&0.8802$_{(0.2013)}$       
   &0.8085$_{(0.0788)}$	&0.6840$_{(0.0954)}$	&0.9271$_{(0.1012)}$\\ 

MedFormer~\cite{wang2024medformer} &  0.7090$_{(0.1610)}$ 	&0.5685$_{(0.1591)}$	&0.8927$_{(0.1935)}$
& 0.8008$_{(0.1013)}$	&0.6747$_{(0.1083)}$	&0.9290$_{(0.0945)}$\\ 

Meng-Net~\cite{meng2021graph} & 0.7876$_{(0.1607)}$	&0.6826$_{(0.1599)}$	&0.8919$_{(0.1845)}$ &0.8233$_{(0.0892)}$	&0.7063$_{(0.1283)}$	 &0.9096$_{(0.1340)}$ \\ 

nnUNet~\cite{isensee2021nnu} & 0.8379$_{(0.1462)}$	&0.7463$_{(0.1568)}$	 &0.8993$_{(0.1679)}$ &0.8814$_{(0.0720)}$	&0.8000$_{(0.1080)}$	&0.8310$_{(0.1112)}$\\ 

D-Persona~\cite{wu2024diversified} & 0.8147$_{(0.1832)}$	&0.7004$_{(0.1386)}$	&0.8431$_{(0.1930)}$ &0.9074$_{(0.0694)}$	&0.8356$_{(0.0924)}$	&0.8963$_{(0.1034)}$\\ 

MAVNet~\cite{yu2024multi} &  0.6343$_{(0.1847)}$	&0.4907$_{(0.1969)}$	&0.5172$_{(0.1902)}$& 0.8026$_{(0.1009)}$	&0.6772$_{(0.1062)}$	&0.6904$_{(0.1048)}$ \\  \hline

MF-Net~\cite{meng2021mf} & 0.7838$_{(0.1801)}$	&0.6742$_{(0.1793)}$	&0.8135$_{(0.1791)}$ & 0.8470$_{(0.0915)}$	&0.7466$_{(0.1193)}$	&0.7743$_{(0.1156)}$ \\

CNV-Net~\cite{vali2023cnv} & 0.7664$_{(0.2097)}$	&0.6514$_{(0.2000)}$	&0.8162$_{(0.2074)}$
&0.8233$_{(0.1084)}$	&0.7063$_{(0.1465)}$	&0.9096$_{(0.1606)}$\\ 

RBGNet~\cite{chen2023rbgnet} &  0.8395$_{(0.1662)}$	&0.7474$_{(0.1768)}$	&0.9020$_{(0.1979)}$& 0.8968$_{(0.0575)}$	&0.8219$_{(0.0895)}$	&0.8713$_{(0.0850)}$ \\  \hline

Proposed   &\textbf{0.8635$_{(0.1746)}$}  & \textbf{0.7902$_{(0.1991)}$}  & \textbf{0.8568$_{(0.1747)}$} & \textbf{0.9189$_{(0.0500)}$}  & \textbf{0.8536$_{(0.0790)}$}  & \textbf{0.9016$_{(0.0808)}$}\\ \bottomrule
\end{tabular}
\label{tab-comparison_test_vessels_Zeiss_Optovue}
\end{table*}
        
\subsection{Comparative analysis on mixed-device datasets}
In this section, we evaluate the performance of MTG-Net on the mixed dataset and compare it with several well-established medical image segmentation methods, including U-Net~\cite{ronneberger2015u}, CE-Net~\cite{gu2019net}, CS-Net~\cite{mou2019cs}, MedFormer~\cite{wang2024medformer}, H2Former~\cite{he2023h2former},TransUNet~\cite{chen2021transunet}, swinUnet~\cite{cao2022swin} and nnUnet~\cite{isensee2021nnu}, as well as the latest general segmentation methods, D-Persona~\cite{wu2024diversified} and MAVNet~\cite{yu2024multi}. Additionally, we benchmark MTG-Net against existing CNV segmentation methods such as MF-Net~\cite{meng2021mf} and RBG-Net~\cite{chen2023rbgnet}. 
To specifically evaluate the effectiveness of the multi-graph interaction-enhanced features designed in MTG-Net, we include a comparison with the method proposed in~\cite{meng2021graph}. Finally, we validate the superior performance of MTG-Net by comparing it with the U$^{2}$Net~\cite{qin2020u2} backbone utilized in our proposed method.

We perform comparative analysis to demonstrate the superior performance of our method across the mixed dataset. Fig.~\ref{fig-compare—image} provides quantitative comparisons of region and vessel segmentation results with respect to different state-of-the-art methods. 
By integrating global and local information through graph neural networks and multi-task learning, MTG-Net shows excellent performance particularly in managing CNV scale variations.
As shown in rows 5-8 in Fig.~\ref{fig-compare—image}, compared to other networks, MTG-Net demonstrates its effectiveness in coping with CNV lesions of various scales. Additionally, as shown in the $2{th}$ and $4{th}$ rows of the vessel segmentation results, as well as the $6{th}$ and $8{th}$ rows of the region segmentation results, the use of uncertainty estimation loss allows MTG-Net to accurately delineate lesion boundaries even in ambiguous regions. 
Unlike other networks, MTG-Net also shows reliability in avoiding over-segmentation when dealing with artifacts and noise that resemble CNV.
Finally, it is important to note that the segmentation results in Fig.~\ref{fig-compare—image} are derived from multiple devices. Despite the varying data conditions across them, MTG-Net consistently shows robust segmentation performance, indicating its superior generalization capability.

Table~\ref{tab-comparison_mix} presents the quantitative results from the comparative analysis. The proposed MTG-Net achieves Dice scores of 87\% for region segmentation and 88\% for vessel segmentation. Compared to classical segmentation methods like TransUNet~\cite{chen2021transunet}, U-Net~\cite{ronneberger2015u}, and CE-Net~\cite{gu2019net}, MTG-Net demonstrates a superior performance, with improvements of 2\%-7\% in region segmentation and 2\%-6\% in vessel segmentation. Notably, TransUNet~\cite{chen2021transunet} provides robust performance due to its effective global dependency obtained from the Transformer architecture, demonstrating its stability across large-scale CNV variations. CE-Net~\cite{gu2019net} performs well in region segmentation through its multi-scale feature extraction and dilated convolution techniques. However, U-Net~\cite{ronneberger2015u} and CS-Net~\cite{mou2019cs} face challenges in accurately delineating irregular CNV shapes and managing large-scale variations. In comparison with CNV-specific networks like CNV-Net~\cite{vali2023cnv} and MF-Net~\cite{meng2021mf}, which struggle with interpreting pixels in fuzzy regions, MTG-Net shows remarkable improvement. It also surpasses RBG-Net by 2\% due to its enhanced network capabilities.
\renewcommand{\tablename}{TABLE}
\begin{table*}[!h]
\setlength{\tabcolsep}{2mm}{} 
\renewcommand\arraystretch{1}
\caption{QUANTITATIVE RESULTS OF ABLATION EXPERIMENTS BASED ON MIXED DATA SETS.}
\centering
\footnotesize
\begin{tabular}{l|cccc|cccc}
\toprule
\multicolumn{1}{c}{\multirow{2}{*}{Methods}} & \multicolumn{4}{c}{Region segmentation} & \multicolumn{4}{c}{Vessel segmentation}\\ 
\cmidrule{2-9}
\multicolumn{1}{c}{~} & DICE  & IoU & Recall &\multicolumn{1}{c|}{Pre}  &  DICE  & IoU & Recall  &\multicolumn{1}{c}{Pre} \\ 
\midrule
M0 & 0.8432$_{(0.1148)}$  & 0.7518$_{(0.1281)}$  & 0.8768$_{(0.1603)}$  & 0.8525$_{(0.1339)}$  & 0.8616$_{(0.0949)}$   & 0.7812$_{(0.1343)}$    & 0.9101$_{(0.1172)}$ & 0.8497$_{(0.1459)}$  \\
\hline
M1 & 0.8489$_{(0.1006)}$   & 0.7522$_{(0.1404)}$   & 0.8824$_{(0.1594)}$  & 0.8566$_{(0.1399)}$  & 0.8660$_{(0.0975)}$   & 0.7734$_{(0.1278)}$    & 0.9034$_{(0.1081)}$ & 0.8411$_{(0.1411)}$  \\
M2 & 0.8534$_{(0.9848)}$   & 0.7591$_{(0.1402)}$   & 0.8846$_{(0.1384)}$  & 0.8647$_{(0.1501)}$  & 0.8698$_{(0.1291)}$   & 0.7831$_{(0.1191)}$    & 0.9045$_{(0.0999)}$ & 0.8489$_{(0.1721)}$  \\
M3 & 0.8623$_{(0.0963)}$ & 0.7654$_{(0.1391)}$ & 0.8867$_{(0.1521)}$ & \textbf{0.8703$_{(0.1431)}$} & 0.8721$_{(0.0942)}$ & 0.7844$_{(0.1293)}$ & 0.9074$_{(0.1061)}$ & 0.8562$_{(0.1234)}$ \\ 
\hline
M$^{*}$1 & 0.8566$_{(0.1181)}$   & 0.7624$_{(0.1374)}$   & 0.8837$_{(0.9753)}$  & 0.8642$_{(0.1425)}$  & 0.8674$_{(0.1156)}$   & 0.7891$_{(0.1335)}$    & 0.9009$_{(0.0982)}$  & 0.8544$_{(0.1405)}$  \\
M$^{*}$2 & 0.8677$_{(0.0918)}$ & 0.7628$_{(0.1285)}$ & 0.8894$_{(0.1432)}$ & 0.8657$_{(0.1346)}$ & 0.8742$_{(0.0836)}$ & 0.7903$_{(0.1281)}$ & 0.9114$_{(0.1289)}$ & 0.8653$_{(0.1349)}$ \\
M$^{*}$3 & \textbf{0.8721$_{(0.0880)}$} & \textbf{0.7740$_{(0.1291)}$} & \textbf{0.9068$_{(0.1050)}$} & 0.8660$_{(0.1401)}$ & \textbf{0.8812$_{(0.0861)}$} & \textbf{0.7974$_{(0.1254)}$} & \textbf{0.9138$_{(0.0933)}$} & \textbf{0.8731$_{(0.1348)}$} \\
\hline
\bottomrule
\end{tabular}
\label{tab-comparison_ablation}
\end{table*}
\subsection{Comparative analysis on single-device datasets}
To evaluate the applicability of MTG-Net across multiple devices, the dataset was divided into device-specific subsets. For each evaluation, one device subset was used as the test set, while the remaining device subsets served as the training set. This ensures a comprehensive segmentation evaluation. Tables~\ref{tab-comparison_test_region_Heidelberg_Svision} and~\ref{tab-comparison_test_region_Zeiss_Optovue} provide the cross-validation quantitative results for region segmentation on individual datasets.
Tables~\ref{tab-comparison_test_vessels_Heidelberg_Svision} and~\ref{tab-comparison_test_vessels_Zeiss_Optovue} provide the cross-validation quantitative results for vessel segmentation on individual datasets. Specifically, when the Heidelberg device is used as the test set, MTG-Net achieves Dice scores of 81.43\% for region segmentation and 81.09\% for vessel segmentation. When the Optovue dataset is used as the test set, the Dice score for region segmentation is 90.38\%, and for vessel segmentation, it also reaches 91.89\%. Furthermore, MTG-Net achieves the best performance on the other two devices. These results illustrate the superior generalizability of our method across multiple devices.

\subsection{Ablation analysis}
In this section, we first use U$^{2}$Net as the backbone network, denoted as M0. Then, we add the edge auxiliary task (T1) on top of M0 to obtain M1 (M0 + T1). Next, we introduce the shape auxiliary task (T2) on top of M1 to obtain M2 (M1+T2). Finally, we incorporate the proposed uncertainty estimation loss component (T3) into M2, resulting in M3 (M2+T3). Meanwhile, under the same experimental settings, we further integrate the MIGR and MRGR modules with M1, M2, and M3, yielding M$^{*}$1, M$^{*}$2, and M$^{*}$3, respectively, to evaluate the effectiveness of each component.

\subsubsection{Ablation for MIGR and MRGR} 
\begin{figure}[!t]
\centering{
\includegraphics[width=7.2cm]{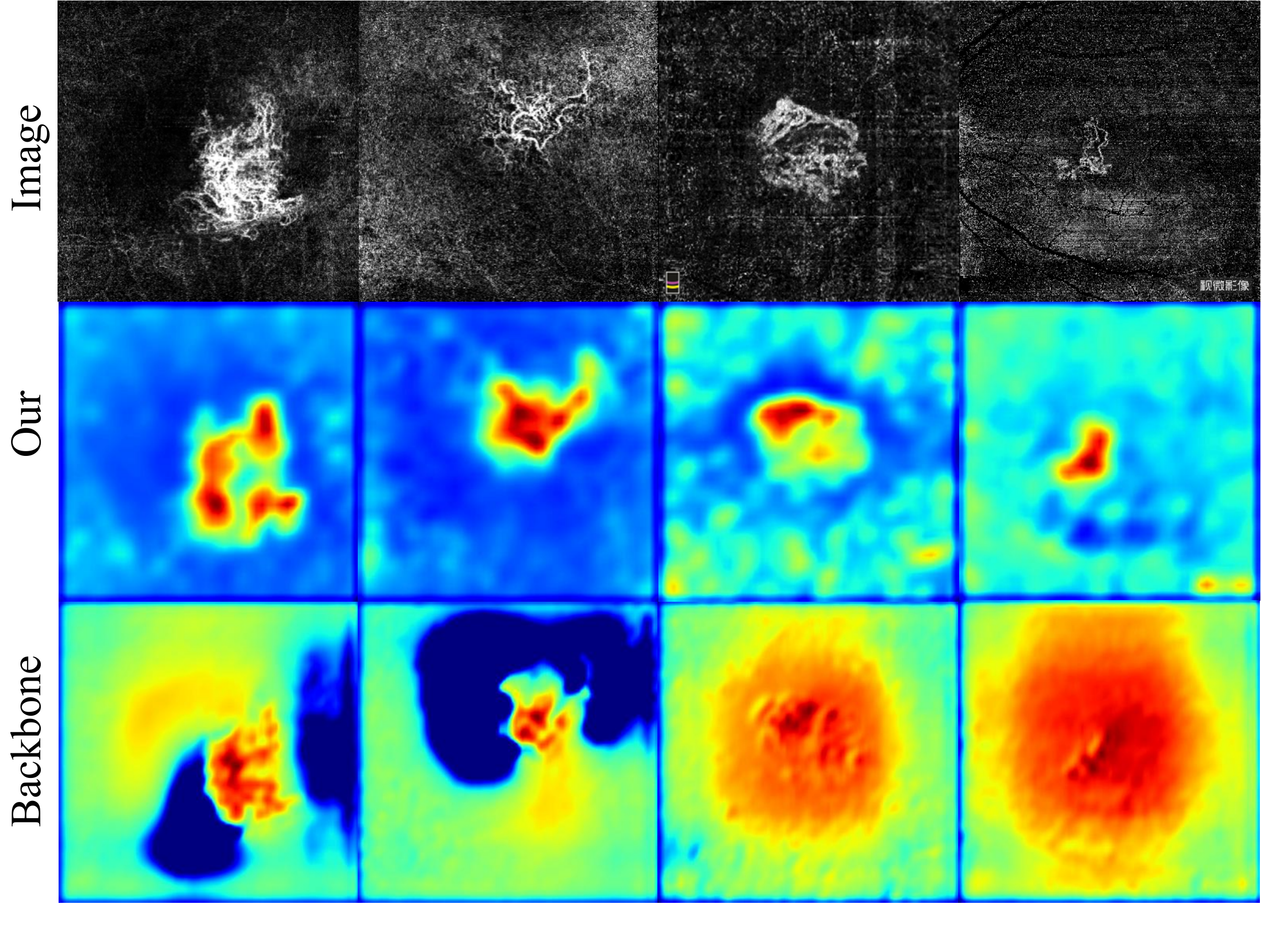}
}
\caption{Comparisons of the attention distributions at the bottleneck layer stage using the MIGR and MRGR-enhanced multitasking network with the multitasking network constituted by the original baseline.}
\label{MIGR}
\end{figure}

Table~\ref{tab-comparison_ablation} shows that the multitask network with MIGR and MRGR outperforms the conventional multitask network, achieving a Dice score of 87\% for region segmentation and 88\% for vessel segmentation. This improvement is mainly due to MIGR and MRGR’s ability to enhance boundary and shape information during training, leading to better generalization and learning. In addition, Fig.~\ref{MIGR} visualizes the bottleneck layer outputs, revealing that our framework localizes lesions and outlines their shapes more effectively than the baseline network. Together, the quantitative and qualitative results confirm the effectiveness of the proposed MIGR and MRGR modules.

\begin{figure}[!t]
\centering{
\includegraphics[width=0.4\textwidth]{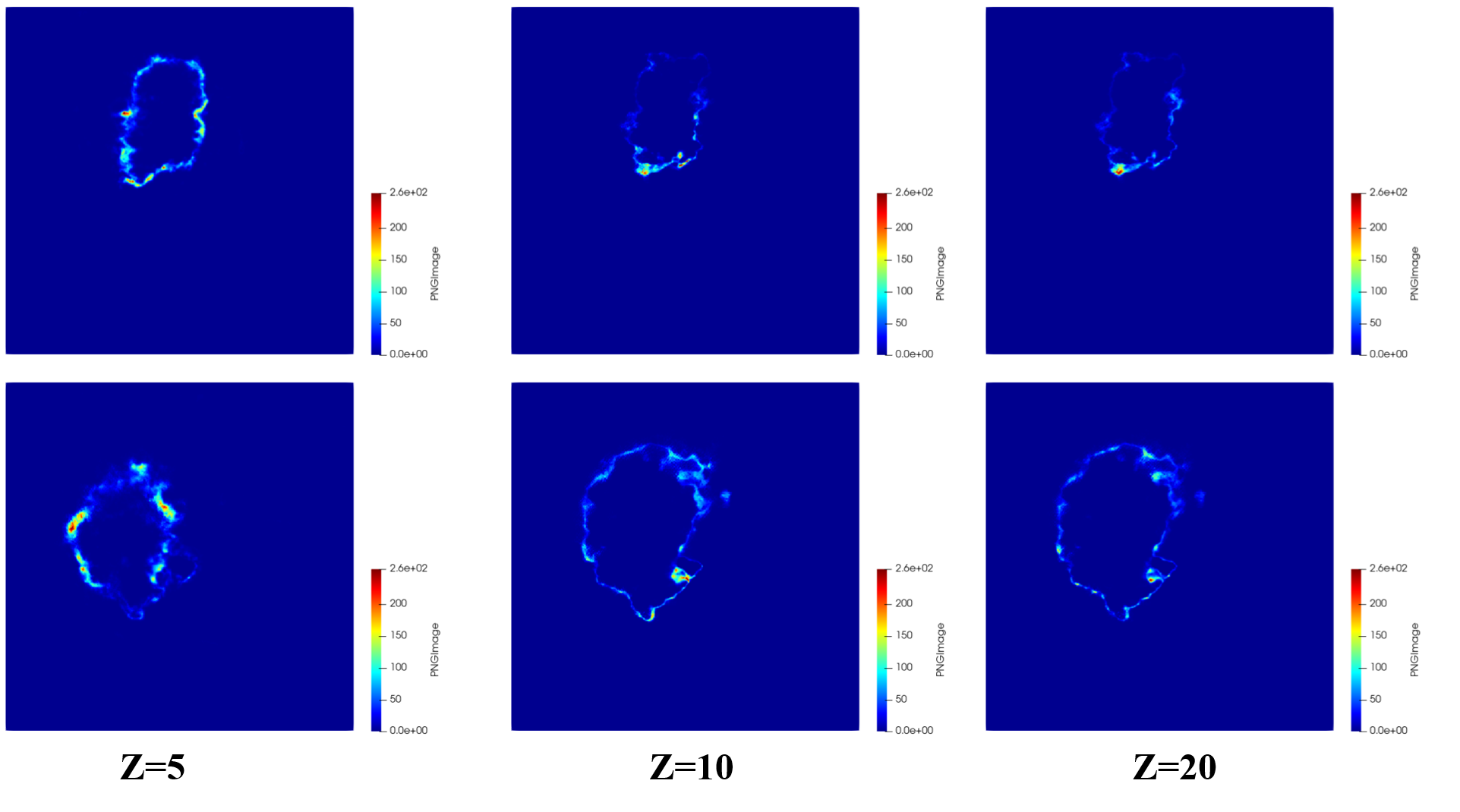}}
\caption{Qualitative analysis of different $Z$.}
\label{Z_sample}
\end{figure}

\begin{figure*}[t]
    \centering{
    \includegraphics[width=15cm]{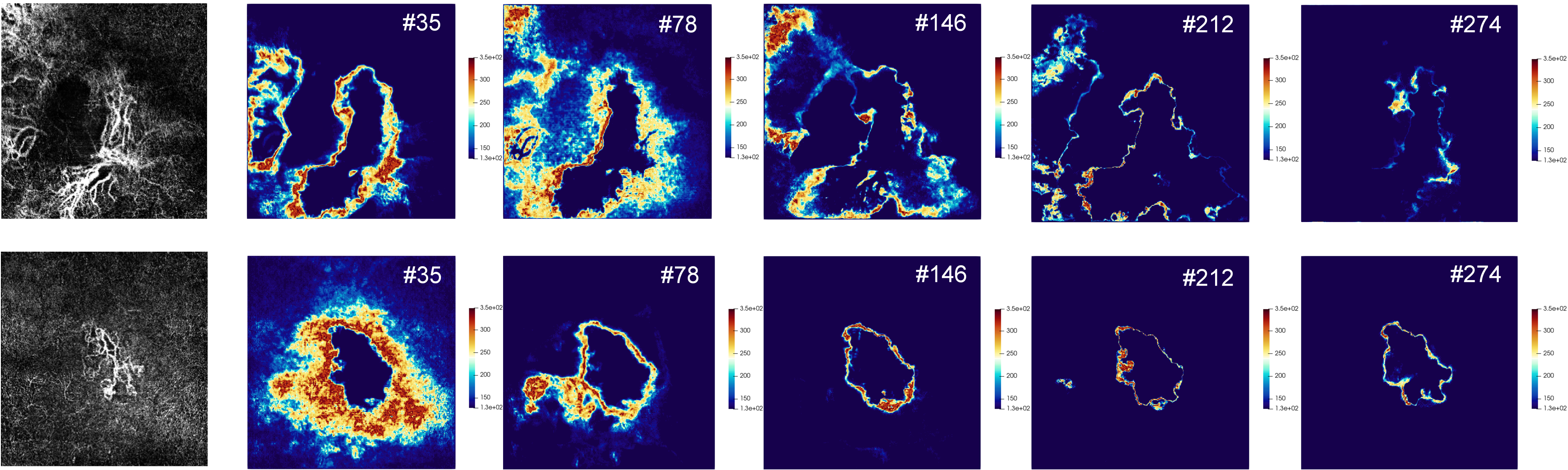}
    }
        \caption{Uncertainty maps during model training. $\#n$ represents the $n$ epoch. During the training phase, the model is constrained by the loss of uncertainty estimation, and the confidence level for dealing with challenging regions gradually increases.}
    \label{uncertainty}
\end{figure*}

\subsubsection{Ablation for uncertainty estimation loss} The uncertainty probability map generated by the proposed uncertainty estimation loss during training is shown in Fig.~\ref{uncertainty}. Here, the symbol $\#n$ denotes the $n_{th}$ training epochs. Our total number of training epochs is 300, and random outputs of the model are selected to illustrate the uncertainty probability changes. From the heat maps in Fig.~\ref{uncertainty}, we can observe that the uncertainty of the CNV segmentation decreases as the number of training epochs increases. This shows the effectiveness of the proposed uncertainty-weighted loss in optimizing fuzzy boundaries.
\begin{figure}[t]
\centering{
\includegraphics[width=8.5cm]{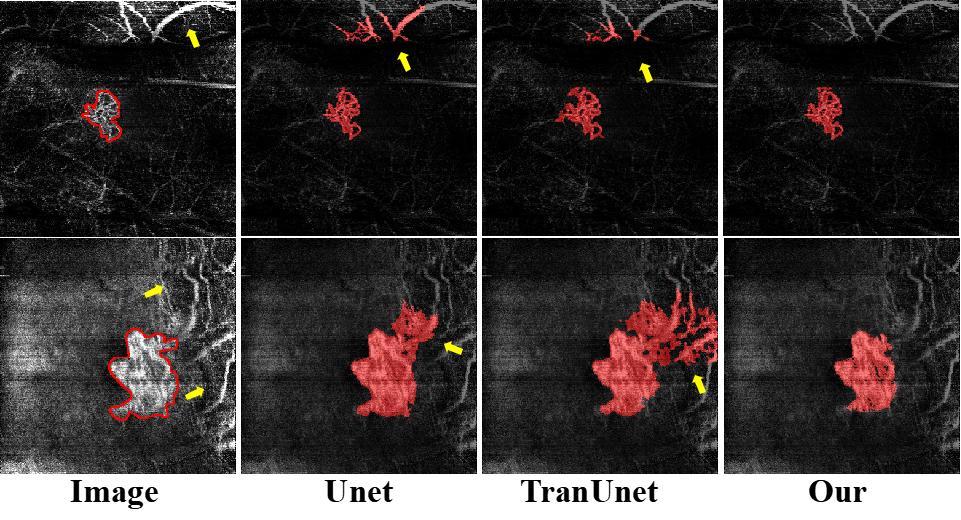}
}
\caption{Comparative analysis results of projection artifacts.}
\label{final}
\end{figure}

\begin{figure*}[t]
    \centering{
    \includegraphics[width=0.9\textwidth]{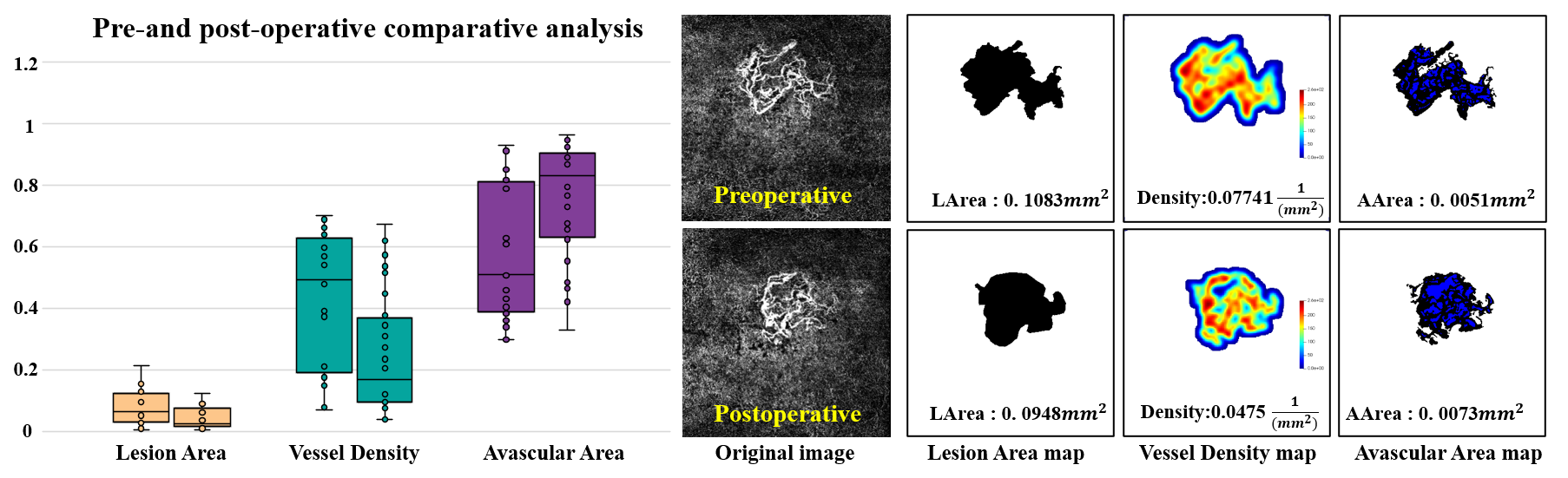}
    }
        \caption{(Left) Paired t-test results based on the area of lesion segmentation, vessel density, and the areas of avascular region segmentation from pre- and post-operations. (Right) Examples of the above measurements based on pre- and post-operations.}
    \label{analysis}
\end{figure*}

\begin{figure}[t]
\centering{
\includegraphics[width=0.48\textwidth]{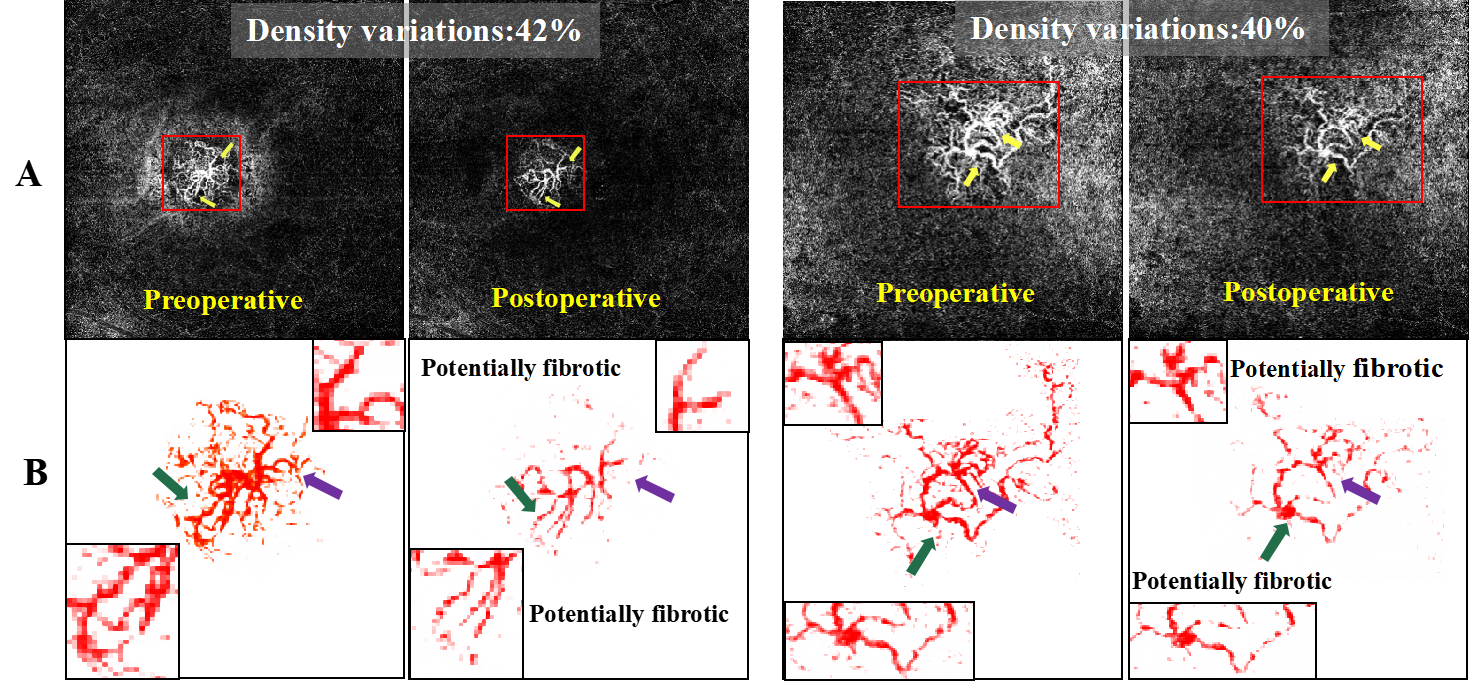}
}
\caption{(A) Comparisons of pre- and postoperative vessel density changes; (B) Visualizations of potentially fibrotic vessels.}
\label{density_compare}
\end{figure}

\begin{table}[!t]
\centering
\caption{PERFORMANCE COMPARISON OF DIFFERENT $K$ VALUES IN THE PROPOSED MODEL. THE TRAINING TIME IS FOR PER EPOCH.}
\label{tab_k_value_performance}
\renewcommand{\arraystretch}{1.25}
\begin{tabular}{c|c|c|c|c}
\toprule
\textbf{K} & \textbf{Dice} & \textbf{IoU} & \textbf{Dimension} & \textbf{Time} \\
\midrule
6 & 0.8621$_{(0.1021)}$ & 0.7667$_{(0.1304)}$ & $6 \times 512$ & 78.46 s\\
12 & \textbf{0.8721$_{(0.0880)}$}  & \textbf{0.7740$_{(0.1291)}$} & $12 \times 512$ & 84.06 s\\
18 & 0.8623$_{(0.0934)}$ & 0.7689$_{(0.1278)}$ & $18 \times 512$ & 94.87 s\\
24 & 0.8767$_{(0.0876)}$ & 0.7774$_{(0.1300)}$ & $24 \times 512$ & 106.24 s\\
\bottomrule
\end{tabular}
\end{table}
\subsection{Parameter sensitivity analysis}
\subsubsection{Effect of node number $K$ in Graph}
The number of graph nodes $K$ is is chosen to balance feature representation complexity and the semantic demands of the specific task. In our approach, we set $K$ to 12. To validate this choice, we tested different $K$ values, as shown in Table~\ref{tab_k_value_performance}. In our framework, the input feature map has a spatial resolution of $12 \times 12$ with 512 channels, setting a theoretical upper limit for $K$ at 144. However, given the low spatial resolution, a smaller value of $K$ is sufficient to capture the feature distribution without unnecessary complexity. Moreover, the number of nodes $K$ is closely related to the model’s computational cost and its sensitivity to noise. While increasing $K$ can capture finer details, it also raises computational demands and makes the model more sensitive to noise—particularly problematic in CNV segmentation, where lesion boundaries are often ambiguous. Too many nodes may amplify this instability and harm performance. Considering the factors above, we set the number of graph nodes $K$ to 12 in our network.
\begin{table}[!t]
\centering
\caption{PERFORMANCE COMPARISON OF DIFFERENT $Z$ VALUES IN THE PROPOSED MODEL. THE TRAINING TIME IS FOR PER EPOCH.}
\label{tab_z_value_performance}
\renewcommand{\arraystretch}{1.25}
\begin{tabular}{c|c|c|c}
\toprule
\textbf{Z} & \textbf{Dice} & \textbf{IoU}  & \textbf{Time} \\
\midrule
5 & 0.8614$_{(0.1221)}$ & 0.7421$_{(0.1344)}$  & 60.79 s \\
10 & \textbf{0.8721$_{(0.0880)}$}  & \textbf{0.7740$_{(0.1291)}$} & 84.06 s\\
20 & 0.8736$_{(0.0812)}$ & 0.7786$_{(0.1278)}$ & 131.57 s\\
\bottomrule
\end{tabular}
\end{table}
\subsubsection{Effect of the number of $Z$ in uncertainty}
Considering that different $Z$-values may have an impact on the training efficiency and segmentation accuracy of the network. Therefore, we experimentally evaluated different sampling values as shown in the table. From Table~\ref{tab_z_value_performance}, it can be seen that $Z=10$ achieves a balance between inference efficiency and segmentation accuracy. In addition, we also performed a visual analysis, as shown in Fig.~\ref{Z_sample}. We observe that when $Z<10$, the uncertainty maps exhibit noticeable fluctuations. While using $Z>10$ does not yield significantly better uncertainty maps compared to $Z=10$, it substantially increases the computational cost.Therefore, setting $Z=10$ is a reasonable choice that ensures stable performance while maintaining computational efficiency.
\section{Discussions}
\subsection{Reliable segmentation against projection artifacts}
As shown in the first column of Fig.~\ref{final} (indicated by the yellow arrow), abnormal CNV tissues in the choroidal layer are easily obscured by projection artifacts from the retinal layers during OCTA imaging. These artifacts can lead to over-segmentation by mistakenly extracting vessel-like structures, as seen in the 2nd and 3rd columns of Fig.~\ref{final}.
In our method, we introduces dual constraints on both boundary and shape to ensure accurate localization of the true lesion boundaries while prevent irregular segmentation caused by artifact interference. Furthermore, the proposed uncertainty loss improves the network's ability to identify regions with high uncertainty caused by projection artifacts. This allows the network to distinguish between reliable predictions and falsely detected structures.
The last column of Fig.~\ref{final} presents the segmentation results of our method, showing its ability to accurately identify lesions while minimizing the impact of artifacts.

\subsection{Pre- and postoperative assessment of wet-AMD}
We performed a dedicate assessment using paired preoperative and postoperative datasets from 26 patients to validate the potential of our method in supporting clinical analysis. As shown on the right side of Fig.~\ref{analysis}, the validation study analyzed three key metrics: lesion area, vessel density, and avascular region area, derived from our segmentation results. The right sub-figure presents the result for each metric. Additionally, we performed paired t-tests on the calculated metrics to statistically evaluate the changes before and after surgery, as depicted on the left side of Fig.~\ref{analysis}. The statistical results indicate that the p-values for all three metrics were less than 0.001, demonstrating a high level of statistical significance. These finding are highly consistent with the actual clinical evidence, confirming the effectiveness our method for providing reliable lesion and vessel segmentations.

\subsection{Assessment of anti-VEGF therapy in wet-AMD}
Currently, the primary clinical treatment for CNV is the injection of anti-vascular endothelial growth factor (anti-VEGF) drugs. Anti-VEGF therapy works by inhibiting the effects of VEGF, thereby reducing the growth and leakage of abnormal blood vessels, and slowing the progression of vision loss. As illustrated in Fig.~\ref{density_compare}(A), which depicts pre- and post-operative outcomes for two patients receiving anti-VEGF therapy, there was an overall reduction in abnormal vessel density of $42\%$ on the left side and $40\%$ on the right side. However, with disease progression and chronic inflammation, neovascularization may become fibrotic, leading to the formation of irreversible scar tissue. Prolonged use of anti-VEGF therapy can sometimes result in regression and scarring of neovascularization, potentially accelerating fibrosis and complicating treatment. As shown in Fig.~\ref{density_compare}(B), although anti-VEGF therapy effectively inhibited the growth of neovascularization, some thicker vessels did not fully regress. This may be due to these vessels being more resistant to treatment or already experiencing fibrosis. Therefore, the quantitative and qualitative analysis from our lesion segmentation can enable clinicians to detect fibrotic tendencies early, helping to prevent further vision loss from fibrosis through more personalized treatment strategies.

\begin{figure}[t]
\centering{
\includegraphics[width=0.5\textwidth]{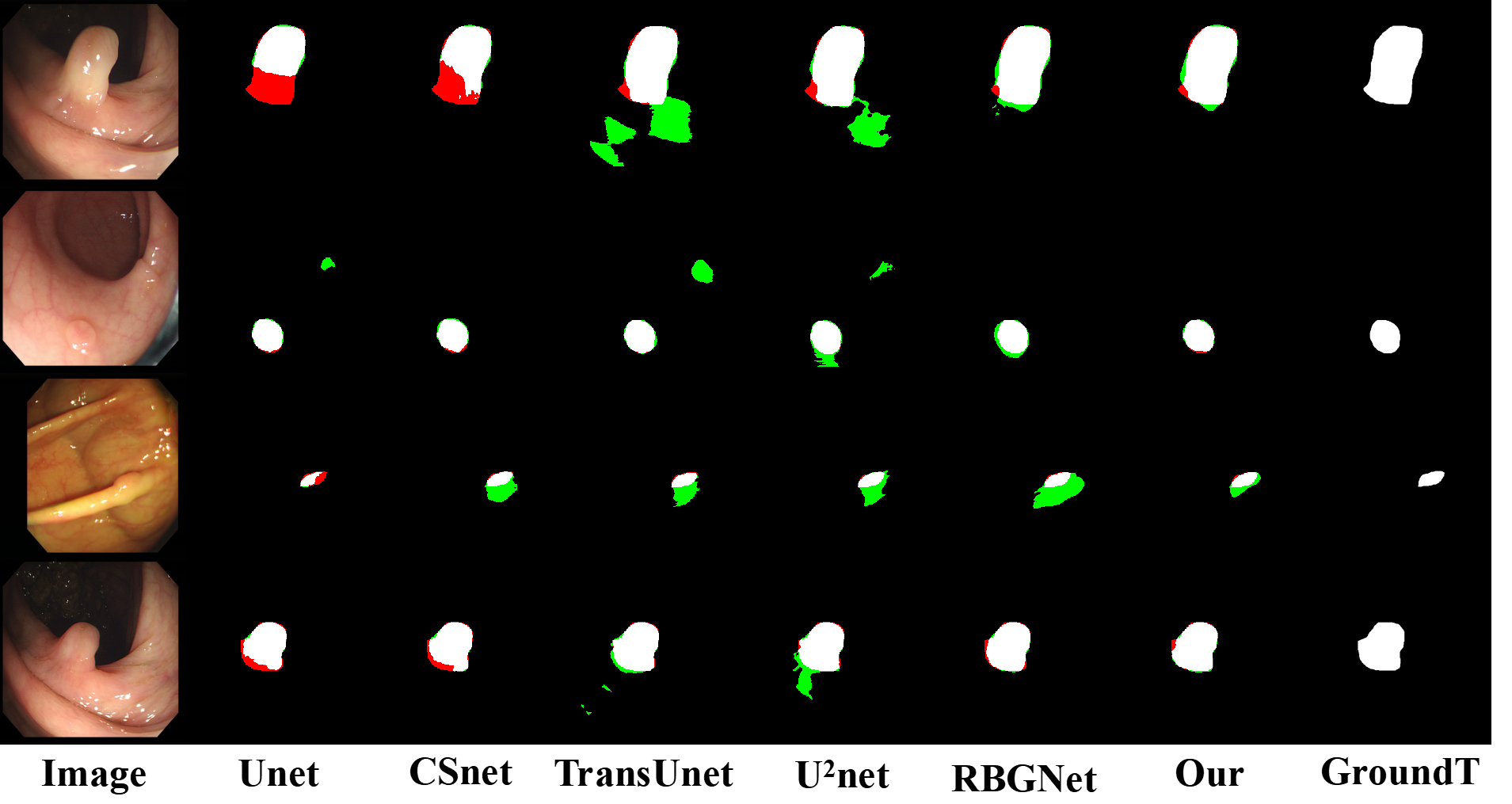}
}
\caption{Comparison of polyp segmentation results. red indicates under-segmentation and green indicates over-segmentation.}
\label{ploy}
\end{figure}

\subsection{Cross-disease validation for generalizability}
To comprehensively evaluate the generalization capability of MTG-Net, we performed extended validation on the benchmark CVC-300 dataset provided by the institution ~\cite{bernal2012towards}, which is used for polyp detection and segmentation in colonoscopy images. The quantitative results, as shown in Table~\ref{tab-poly}, show that our MTG-Net obtains a Dice score of 89\% and an IoU of 84\%, both achieving the best performance across all comparisons. The qualitative results, illustrated in Fig.~\ref{ploy}, further illustrating that MTG-Net is able to accurately delineate polyp boundaries and achieve precise lesion segmentation performance, particularly when handling polyp data with blurry boundaries. This indicates the potential of using MTG-Net for a wide range of applications.
\renewcommand{\tablename}{TABLE}
\begin{table}[!h]
\setlength{\tabcolsep}{1mm}{} 
\renewcommand\arraystretch{1}
\caption{COMPARATIVE EXPERIMENTAL RESULTS OF POLYP SEGMENTATION IN CVC-300 DATASETS.}
\centering
\footnotesize
\begin{tabular}{l|ccccc}
\toprule

\multicolumn{1}{c}{\multirow{2}{*}{Methods}} & \multicolumn{5}{c}{Region segmentation} \\ 
\cmidrule{2-6}

\multicolumn{1}{c}{~} & DICE  & IoU & Recall & FDR   &\multicolumn{1}{c}{Pre}  \\ 
\midrule

U-Net~\cite{ronneberger2015u} & 0.8154	&0.7489	&0.8885	&0.1731	&0.8102	\\
CE-Net~\cite{gu2019net} & 0.8458	&0.7746	&0.9216	&0.1690	&0.8310  \\
CS-Net~\cite{mou2019cs} & 0.8273	&0.7468	&0.9335	&0.2174	&0.7826	  \\
TransUNet~\cite{chen2021transunet} &0.7018	&0.6022	&0.8300	&0.3261	&0.6739 \\ 
SwinUnet~\cite{cao2022swin} & 0.7316	&0.6212	 &0.8465	&0.1934	 &0.7524 \\ 
U$^{2}$Net~\cite{qin2020u2} & 0.8348	&0.7639	&0.9064	&0.1954	&0.8046 \\ 
H2Formert~\cite{he2023h2former} & 0.8059   &0.7247   & 0.8458 & 0.1648  &0.8185  \\ 
MedFormer~\cite{wang2024medformer} & 0.8208	&0.7283	 &0.9146	&0.2114	&0.7885\\ 
Meng-Net~\cite{meng2021graph} & 0.8410  & 0.7727  & 0.9462 & 0.2010  &0.7990  \\ 
nnUNet~\cite{isensee2021nnu} & 0.8725	 &0.8007	&0.9226	 &0.1461	&0.8538  \\ \hline
MF-Net~\cite{meng2021mf} & 0.8552  & 0.7860   & 0.9436  & 0.1749& 0.8250  \\
CNV-Net~\cite{vali2023cnv} & 0.8214  & 0.7312  & 0.8567 & 0.1727  &0.8234 \\ 
RBGNet~\cite{chen2023rbgnet} & 0.8701  & 0.8167  & 0.9482 & 0.1519  &0.8492  \\ \hline
Proposed   & \textbf{0.8979}  & \textbf{0.8489}   & \textbf{0.9527} & \textbf{0.1128} & \textbf{0.8872}\\ \bottomrule
\end{tabular}
\label{tab-poly}
\end{table}

\subsection{Limitations}
We propose a novel method for CNV segmentation based on OCTA images. However, several limitations remain. First, the model's computational efficiency poses a significant challenge. Although the nested U-shaped framework and the integration of uncertainty assessment loss improve the network’s ability to capture multi-scale information and resolve ambiguous boundaries, these enhancements come with a notable increase in computational demands. Despite the use of dilated convolutions to reduce parameter complexity and streamline processing, the model’s structural depth still results in substantial computational costs. Thus, knowledge distillation techniques could be explored to transfer knowledge from a larger teacher model to a more lightweight student model without altering the network architecture, thereby reducing computational overhead and enhancing training efficiency.
Additionally, although the proposed method has been validated across different OCTA devices, the data from devices other than Heidelberg are limited. This lack of variability may reduce the model's generalizatoin ability. Expanding the dataset to incorporate more diverse data sources would help mitigate this limitation and enhance the model's robustness.

\section{Conclusions}
In this study, we propose MTG-Net, an interaction-enhanced learning framework based on graph convolutional networks for neovascularization segmentation in OCTA images. Our model integrates two novel modules, MIGR and MRGR, which work together to extract key information from CNV lesions, improving boundary clarity and shape recognition. We formulate MTG-Net as a recurrent graph reasoning process to better organize and refine morphological features during training. Additionally, we incorporate uncertainty estimation to enhance the model’s ability to handle blurry boundaries. Extensive experiments on four clinical CNV datasets and one polyp dataset demonstrate the effectiveness and generalizability of our method, showing its potential for related tasks.

\bibliographystyle{IEEEbib}
\bibliography{refs.bib}

\vfill
\end{document}